\documentclass{article}
\usepackage{ijcai21}

\usepackage{times}
\usepackage{soul}
\usepackage{url}
\usepackage[hidelinks]{hyperref}
\usepackage[utf8]{inputenc}
\usepackage[small]{caption}
\usepackage{graphicx}
\usepackage{amsmath}
\usepackage{amsthm}
\usepackage{booktabs}
\usepackage{algorithm}
\usepackage{algorithmic}

\usepackage{subfigure}
\usepackage{wrapfig}
\usepackage[switch]{lineno}
\usepackage{color}
\usepackage{todo}
\usepackage{multirow}

\title{Transfer with Action Embeddings for Deep Reinforcement Learning}

\author{
     Yu Chen$^1$ \thanks{Equal contribution} \and 
     Yingfeng Chen$^{1*}$\and
     Zhipeng Hu$^2$\and
     Tianpei Yang$^3$\and \\
     Changjie Fan$^1$\and
     Yang Yu$^4$\and
     Jianye Hao$^3$
     \affiliations
     $^1$ NetEase Fuxi AI Lab\\
     $^2$ Zhejiang University\\
     $^3$ College of Intelligence and Computing, Tianjin University\\
     $^4$ National Key Laboratory for Novel Software Technology, Nanjing University
     \emails
     \{chenyu5,chenyingfeng1,fanchangjie\}@corp.netease.com,
     11921156@zju.edu.cn,
     \{tpyang,jianye.hao\}@tju.edu.cn,
     yyu@lamda.nju.edu.cn
}
\begin{document}
% \linenumbers
\maketitle

\begin{abstract}
     % Despite the great success achieved in various sequential decision tasks, deep reinforcement learning is still criticized for its data inefficiency.
     % Many approaches have been proposed to alleviate this issue, and transfer learning is a promising direction.
     Transfer learning (TL) is a promising way to improve the sample efficiency of reinforcement learning.
     % Most of the previous work focuses on transferring between tasks with the same state-action space. 
     However, how to efficiently transfer knowledge across tasks with different state-action spaces is investigated at an early stage.
     % In this paper, we propose a novel method to facilitate efficient transfer across different state-action domains. Specifically, we first learn action embeddings by extracting similar action semantics between tasks. Then we directly transfer the learned policy to the target task.
     Most previous studies only addressed the inconsistency across different state spaces by learning a common feature space, without considering that similar actions in different action spaces of related tasks share similar semantics. In this paper, we propose a method to learning action embeddings by leveraging this idea, and a framework that learns both state embeddings and action embeddings to transfer policy across tasks with different state and action spaces.
     %% We evaluate our methods through a series of tasks, including gridworld navigation tasks, discretized continuous control tasks, and combat tasks in a commercial game. 
     Our experimental results on various tasks show that the proposed method can not only learn informative action embeddings but accelerate policy learning.

\end{abstract}

\section{Introduction}
Deep reinforcement learning (DRL), which combines reinforcement learning algorithms and deep neural networks, has achieved great success in many domains, such as playing Atari games \cite{mnih2015human}, playing the game of Go \cite{silver2016mastering} and robotics control \cite{levine2016end}.
Although DRL is viewed as one of the most potential ways to General Artificial Intelligence (GAI), it is still criticized for its data inefficiency. Training an agent from scratch requires considerable numbers of interactions with the environment for a specific task.
% Moreover, some researchers point out that DRL algorithms are likely to overfit to the trained environment \cite{banerjee2018deep}. Once the configuration of the environment changes, the learned policy will not work and needs to be retrained. 
% How to utilize the knowledge learned from previous tasks in a new related task remains a challenge.
One approach dealing with this problem is \textit{Transfer Learning} (TL) \cite{taylor2009transfer}, which makes use of prior knowledge from relevant tasks to reduce the consumption of samples and improve the performance of the target task.

Most of the current TL methods are essential to share the knowledge that is contained in the parameters of neural networks. However, this kind of knowledge cannot be directly transferred when faced with the cross-domain setting, where the source and target tasks have different state-action spaces. In order to overcome the domain gap, some researches focus on mapping state spaces into a common feature space, such as using manifold alignment \cite{ammar2015unsupervised,gupta2017learning}, mutual information \cite{wan2020mutual} and domain adaptation \cite{carr2019domain}.
%These methods leverage state embeddings for transfer, but either ignore structure in the space of actions. 
However, none of them considers similar actions in different action spaces of related tasks share similar semantics.
%Different from them, this work puts particular emphasis on the action spaces, and we claim that action spaces of related tasks also share some semantic structure.
% Action representations have been studied recently and used to improve function approximation \cite{TennenholtzM19}, generalization \cite{ChandakTKJT19,jain2020generalization} or transfer between tasks with the same state-action spaces \cite{whitney2019dynamicsaware}.
% Different from them, this work aims to utilize action representations to transfer between tasks with different state-action spaces
% we claim that action spaces of related tasks also share some semantic structure.
To illustrate this insight, we take the massively multiplayer online role-playing game (MMORPG) as an example. An MMORPG usually consists of a variety of roles, each of which is equipped with a set of unique skills. However, these skills share some similarities since some of them cause similar effects, such as 'Damage Skill', 'Control Skill', 'Evasion Skill' and so on. %online games as an example. %In many online games, different classes have their unique skills, but the effects of skills may be similar. We usually can divide their skills into several categories, such as 'Damage Skill', 'Control Skill', 'Evasion Skill' and so on.
% Recent work have extracted action representations from expert demonstrations \cite{TennenholtzM19} or without prior knowledge \cite{ChandakTKJT19}. And it is used to improve function approximation \cite{TennenholtzM19}, generalization \cite{ChandakTKJT19,jain2020generalization} or transfer between tasks with the same state-action spaces \cite{whitney2019dynamicsaware}.
% Inspired by these observations and recent work on action embedding \cite{TennenholtzM19,ChandakTKJT19}, we study the feasibility of leveraging action embeddings, which automatically learn the semantics of actions, to transfer policy across tasks with different state spaces and/or action spaces. 
Several work have studied action representations and used them to improve function approximation \cite{TennenholtzM19}, generalization \cite{ChandakTKJT19,jain2020generalization} or transfer between tasks with the same state-action spaces \cite{whitney2019dynamicsaware}.
In contrast to these researches, we study the feasibility of leveraging action embeddings to transfer policy across tasks with different action spaces. Intuitively, similar actions will be taken when performing tasks with the same goal. Hence, if the semantics of actions is learned explicitly, there is a chance to utilize the semantic information to transfer policy.
% add a figure to illustrate the idea

The main challenge in the action-based transfer is how to learn meaningful action embeddings that captures the semantics of actions. Our key insight is that the semantics of actions can be reflected by their effects which are defined by state transitions in RL problems. Thus, we learn action embeddings through a transition model, which predicts the next state according to the current state and action embedding. % , without any prior knowledge
Another challenge is that how to transfer policy across tasks with different state and action spaces.
To this end, we propose a novel framework named TRansfer with ACtion Embeddings (TRACE) for DRL, where we leverage both state embeddings and action embeddings for policy transfer.
When transferring to the target task, we transfer both the policy model and the transition model learned from the source task. The nearest neighbor algorithm used by the policy model can select the most similar action in the embedding space. Meanwhile, the transition model helps to align action embeddings of the two tasks.
The main contributions are summarized as follows:
%Our main contributions in this paper are summarized as follows: 
\begin{itemize}
     \item We propose a method to learn action embeddings, which can capture the semantics of the actions.
           %which may be used for other purposes, such as action clustering \cite{cuayahuitl2019deep}.

           %\item The learned action embeddings help policy transfer by using the nearest neighbor algorithm.

     \item We propose a novel framework TRACE which learns both state embeddings and action embeddings to transfer policy across tasks with different state and action spaces.
           % \item We propose a transfer framework via learned action embeddings to transfer policy across tasks with different state and/or action spaces.

     \item Our experimental results show that TRACE can
           a) learn informative action embeddings;
           b) effectively improve sample efficiency compared with vanilla DRL and state-of-the-art transfer algorithms.
           % \footnote{The experimental code of this work is released anonymously at \url{https://github.com/ActionEmbedding/ActionEmbedding.git} for reproducibility. } 
\end{itemize}
%Specifically, 
% where RL methods are combined with action embeddings by using the nearest neighbor algorithm.

% When transfer policy to the target task, we transfer both the policy model and the transition model learned from the source task.
% By reusing the transition model, actions of the source and target tasks are embedded into the same or similar space, helping the policy adapt to the target task quickly.
%Moreover, our experiments conducted on three sets of tasks show that our method can effectively transfer the policy across similar tasks.

%Another difficulty is that the embeddings of the source and the target tasks need to be aligned such that the policy from the source task can still take similar actions in the target task. In this work, we address the issue by fixing the transition model and updating action embeddings, which ensures that the actions of source and target tasks are embedded into the same or similar space. \textcolor{red}{modify the second difficulty}

%To evaluate TRACE empirically, we test the method with Soft Actor-Critic (SAC) \cite{haarnoja2018soft} on three sets of tasks: gridworld navigation tasks, discretized continuous control tasks, and combat tasks in a commercial game. The experimental results show that TRACE significantly improves the speed of subsequent learning. 

\section{Related Work}
\subsection{Transfer in Reinforcement Learning}
Transfer learning is considered an important and challenging direction in reinforcement learning and has become increasingly important.
\cite{teh2017distral} proposes a method that uses a shared distilled policy for joint training of multiple tasks named Distral.
\cite{finn2017model} introduces a general framework of meta-learning that can achieve fast adaptation.
Successor features and generalized policy improvement are also applied to transfer knowledge \cite{barreto2019transfer,ma2018universal}.
\cite{yang2020efficient} proposes a Policy Transfer Framework (PTF) which can effectively select the optimal source policy and accelerate learning.
All these methods focus on the tasks that only differ in reward functions or transition functions.
% or transition functions \cite{yang2020efficient}.

To transfer across tasks with different state and action spaces, many research attempts to map state spaces into a common feature space.
\cite{gupta2017learning} learns common invariant features between different agents from a proxy skill and uses it as an auxiliary reward. However, it requires corresponding state pairs of two tasks, which may be difficult to acquire.
Adversarial training \cite{wulfmeier2017mutual,carr2019domain} is also utilized to align the representations of source and target tasks.
Additionally, MIKT \cite{wan2020mutual} learns an embedding space with the same dimension as the source task and uses lateral connections to transfer knowledge from teacher networks, which is similar to \cite{liu2019knowledge}.
These methods focus on the connection between state spaces but ignore the relationship between action spaces.
% \textit{Adversarial losses}, which are based on the mutual alignment of visited state distributions between tasks, are used by \cite{wulfmeier2017mutual} as auxiliary rewards for training policies.
% \cite{carr2019domain} later adopts Adversarial Autoencoder (AAE) to align the representation vectors of target and source states on atari games.
Recently, OpenAI Five \cite{raiman2019neural} introduces a technique named Surgery to determine which section of the model should be retrained when the architecture changes. The method is capable of continuous learning but is not suitable for tasks with totally different action spaces.

Inter-task mapping, which considers both state and action spaces, is used to describe the relationship between tasks through explicit task mappings.
\cite{taylor2007transfer} manually constructs an inter-task mapping and builds a transferable action-value function based on it.
% In \cite{ammar2011reinforcement}, they introduce a common task subspace between states of tasks and use it to learn the state mapping between tasks.
Furthermore, Usupervised Manifold Alignment (UMA) is used to learn an inter-task mapping from trajectories of source and target tasks autonomously \cite{ammar2015unsupervised}.
\cite{zhang2021learning} learns state and action correspondence across domains using a cycle-consistency constraint to achieve policy transfer.
The main difference from our work is that we try to embed actions into a common space instead of learning a direct mapping.

% While in this paper, we investigate the representations of discrete actions and use them to improve the RL algorithms.

\subsection{Action Embedding}
Action embedding is firstly studied by \cite{dulac2015deep}, aiming to solve the explosion of action space in RL.
However, the action embeddings are assumed to be given as a prior.
Act2Vec is introduced by \cite{TennenholtzM19}, in which a skip-gram model is used to learn action representations from expert demonstrations.
% And they transfer the embeddings from a 2D navigation task to a 3D navigation task.
\cite{ChandakTKJT19} learns a latent space of actions by modeling the inverse dynamics of the environment, and a function is learned to map the embeddings into discrete actions.
While in this paper, we learn representations of discrete actions through a forward transition model.
Similarly, \cite{whitney2019dynamicsaware} simultaneously learns embeddings of states and action sequences that capture the environment's dynamic to improve sample efficiency.
% Further, they show that the learned embeddings can generalize to other tasks in the same domain. 
However, they focus on continuous control tasks and consider the effects of action sequences.
Action representations are also used to enable generalization to unseen actions \cite{jain2020generalization}. By contrast, we use it to transfer policy across different tasks.
% jointly learn state and action embeddings.

% \subsection{Representation Learning}
% Representation learning aims to learn representations of the data that make it easier to extract useful
% information when building classifiers or other predictors \cite{bengio2013representation}, and have been applied in various domains, e.g., NLP \cite{mikolov2013efficient,peters2018deep} and graphs \cite{nie2017unsupervised}.
% In reinforcement learning, features are extracted from raw images by convolutional neural networks \cite{mnih2015human}, sparse representations are learned for control \cite{liu2019utility}, and the work \cite{franccois2019combined} combine model-based and model-free approaches via a shared state abstraction.
% Moreover, in \cite{hausman2018learning}, they utilize skill representations to learn versatile skills in hierarchical reinforcement learning.
% While in this paper, we investigate the representations of discrete actions and use it to improve the RL algorithms.

% \section{Problem Formulation and Assumptions}
\section{Problem Definition}
RL problems are often modeled as Markov Decision Processes (MDPs) which are defined as a tuple $\mathcal{M} = (\mathcal{S}, \mathcal{A}, \mathcal{T}, \mathcal{R}, \gamma)$, where $\mathcal{S}$ and $\mathcal{A}$ are sets of states and actions. In this work, we restrict our method on discrete action spaces, and $|\mathcal{A}|$ denotes the size of action set.
$\mathcal{T}: \mathcal{S} \times \mathcal{A} \times \mathcal{S} \mapsto [0, 1]$ is a state transition probability function, which can also be represented as the distribution of resulting states $p(s_{t + 1}|s_t, a_t)$ at time step $t$.
$\mathcal{R}: \mathcal{S} \times \mathcal{A} \mapsto \mathbb{R}$ is a reward function measuring the performance of agents and $\gamma$ is a discount factor for future rewards.
Additionally a policy $\pi: \mathcal{S} \times \mathcal{A} \mapsto [0, 1]$ is defined as a conditional distribution over actions for each state.
% For any policy $\pi$, its corresponding state value function is $V_\pi (s) = \mathbb{E}[\sum_{i=0}^\infty \gamma^i r_{t+1} | s_t = s]$ and state-action value function is $Q_\pi(s, a) = \mathbb{E}[\sum_{i=0}^\infty \gamma^i r_{t+1} | s_t = s, a_t = a]$ for all $s \in \mathcal{S}$ and $a \in \mathcal{A}$ at time step $t$.
Given an MDP $\mathcal{M}$, the goal of the agent is to find an optimal policy $\pi^*$ that maximizes the expected discounted return $R = \sum_{t=0}^{\infty} \gamma^t r_{t}$.

In this paper, we consider the transfer problem between a source MDP $\mathcal{M}_S = (\mathcal{S}_S, \mathcal{A}_S, \mathcal{T}_S, \mathcal{R}_S, \gamma_S)$ and a target MDP $\mathcal{M}_T = (\mathcal{S}_T, \mathcal{A}_T, \mathcal{T}_T, \mathcal{R}_T, \gamma_T)$.
% Generally, the state and action spaces in the two MDPs are different, so are the dynamics $\mathcal{T}$ and reward function $\mathcal{R}$.
% However, in this paper, we assume that there are some similarities in both reward functions and transition functions. 
In this paper, we assume that the state and action spaces in the two MDPs are different, while there are some similarities in both the reward functions and the transition functions.
For example, in one of our experimental tasks, $\mathcal{M}_S$ and $\mathcal{M}_T$ correspond to two different roles in a MMORPG to fight against an enemy. While the dimensions of actions (skills) and states are completely different, the two agents both need to defeat the enemy, with a reward that depends on the final result: win, lose or tie. Besides, though the actions (skills) of the two roles are different, their effects can be similar, such as two agents both choose a damage skill which cause similar damage to the enemy. %Therefore, the reward and the transition functions share some similarities between the two task. 
% In particular, the goals(rewards) of the two MDPs are similar so that the optimal policy of the target MDP will resemble the optimal policy of the source MDP. Meanwhile, we may embed the tasks' action into the same or similar semantic space due to transition similarity.
% \textcolor{red}{assume actions of two MDPs have the same semantic space?}

% \section{Methods} % The name of the method
\section{Transfer with Action Embeddings}
In this section, we introduce the TRACE framework.
We first discuss how to learn meaningful action embeddings.
Further, we describe how the action embeddings can be combined with RL algorithms and to facilitate policy transfer.

\subsection{Learning Action Embeddings}
To learn action embeddings that capture the semantics of actions, our main insight is that the semantics of actions can be reflected by their effects on the environment, which can be measured by the state transition probability in RL.
% Thus, we aim to learn an action embedding $e(a) \in \mathbb{R}^d$ for each $a \in \mathcal{A}$, such that the distance between action embeddings , whose effect on the environment is similar, is minimized.
Thus, we aim to learn an action embedding $e(a) \in \mathbb{R}^d$ with dimension $d$ for each $a \in \mathcal{A}$, which should satisfy some properties: (1) The distance between action embeddings is adjacent if the actions have similar effects on the environment. (2)  The embeddings should be sufficient so that the distribution conditioned on the embeddings approximates that conditioned on the actions $p(s_{t+1}|s_t, a_t) \approx p(s_{t+1}|s_t, e(a_t))$.
We approximate this by learning a transition model $f_{\theta^D}$, which predicts the next state $\tilde{s}_{t + 1}$ according to current state $s_t$ and action $a_t$ with parameter $\theta^D$, to capture the dynamics of the environment.

Figure \ref{fig: onestep} illustrates the learning process of action embeddings. At the beginning, an embedding matrix $W^{ae} \in \mathbb{R}^{|\mathcal{A}| \times d}$ is instantiated, in which the $i$-th row of the matrix denotes the embedding vector $e(a_i)$.
Given a state transition tuple $(s_t, a_t, s_{t+1})$, we first get action embedding $e(a_t)$ by lookup from the current matrix $W^{ae}$. Then a latent variable $z_t$ is sampled from $z_t \sim \mathcal{N}(\mu_{t}, \sigma_{t})$, where $[\mu_t, \sigma_t] = f_{\theta^{D_1}}(s_t, e(a_t))$, like variational autoencoder (VAE) \cite{kingma2013auto}.
The latent variable is introduced as a stochastic process to cope with stochastic environments, which is similar to \cite{goyal2017z}.
Further, the model predicts the next state $\tilde{s}_{t+1}$ by a multi-layered feed-forward neural network that conditions on $z_t$, $s_t$, and $e(a_t)$, i.e. $\tilde{s}_{t+1} = f_{\theta^{D_2}}(s_t, e(a_t), z_t)$.
The transition model and action embeddings are optimized to minimize the prediction error:

\begin{figure}
     \centering
     \includegraphics[height=100pt]{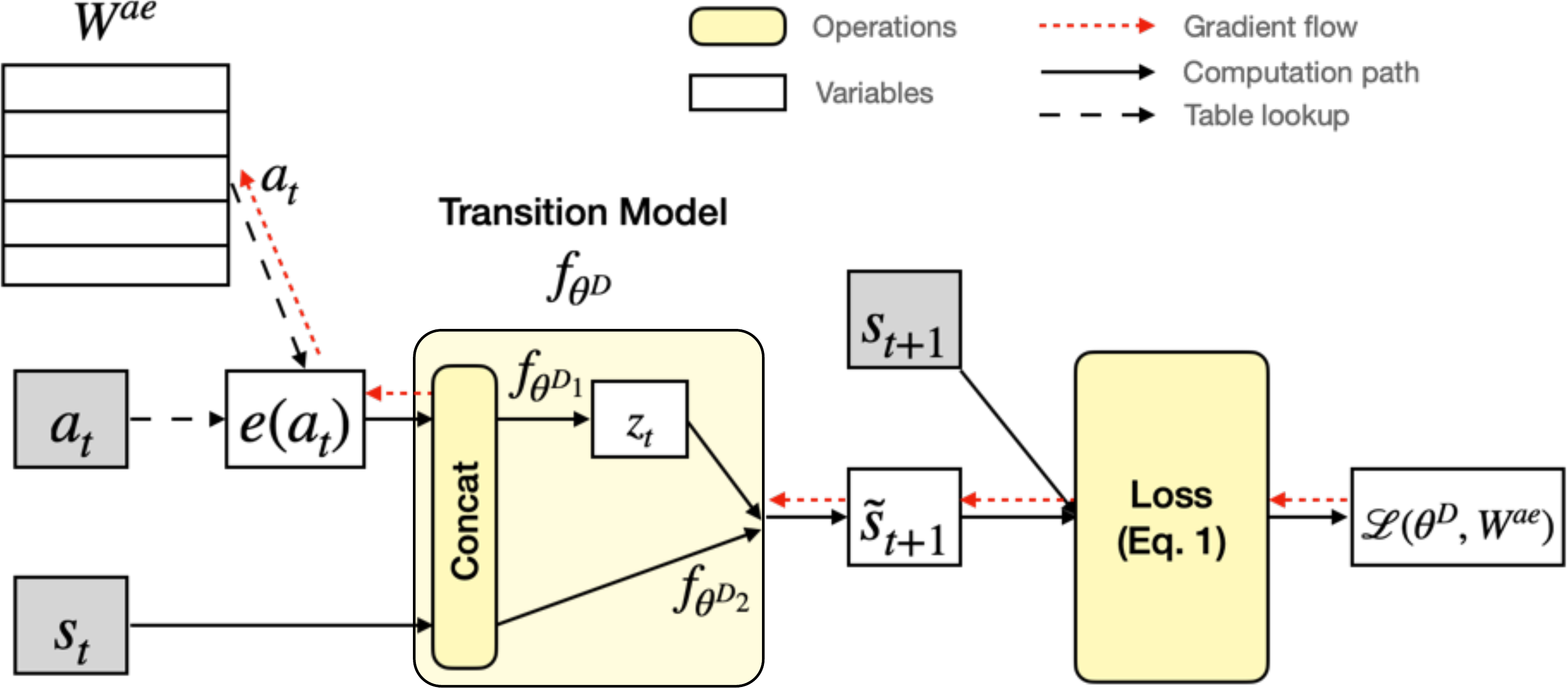}
     \caption{Illustration of the process of learning action embeddings. Given a state transition tuple, we first obtain action embedding $e(a_t)$ from $W_{ae}$. The transition model is used to predict the next state $\tilde{s}_{t+1}$. The red arrow denotes the gradients of the loss, and the grey grid denotes input variables.}
     \label{fig: onestep}
     \vspace{-10pt}
\end{figure}

\begin{figure*}
     \centering
     \includegraphics[height=160pt]{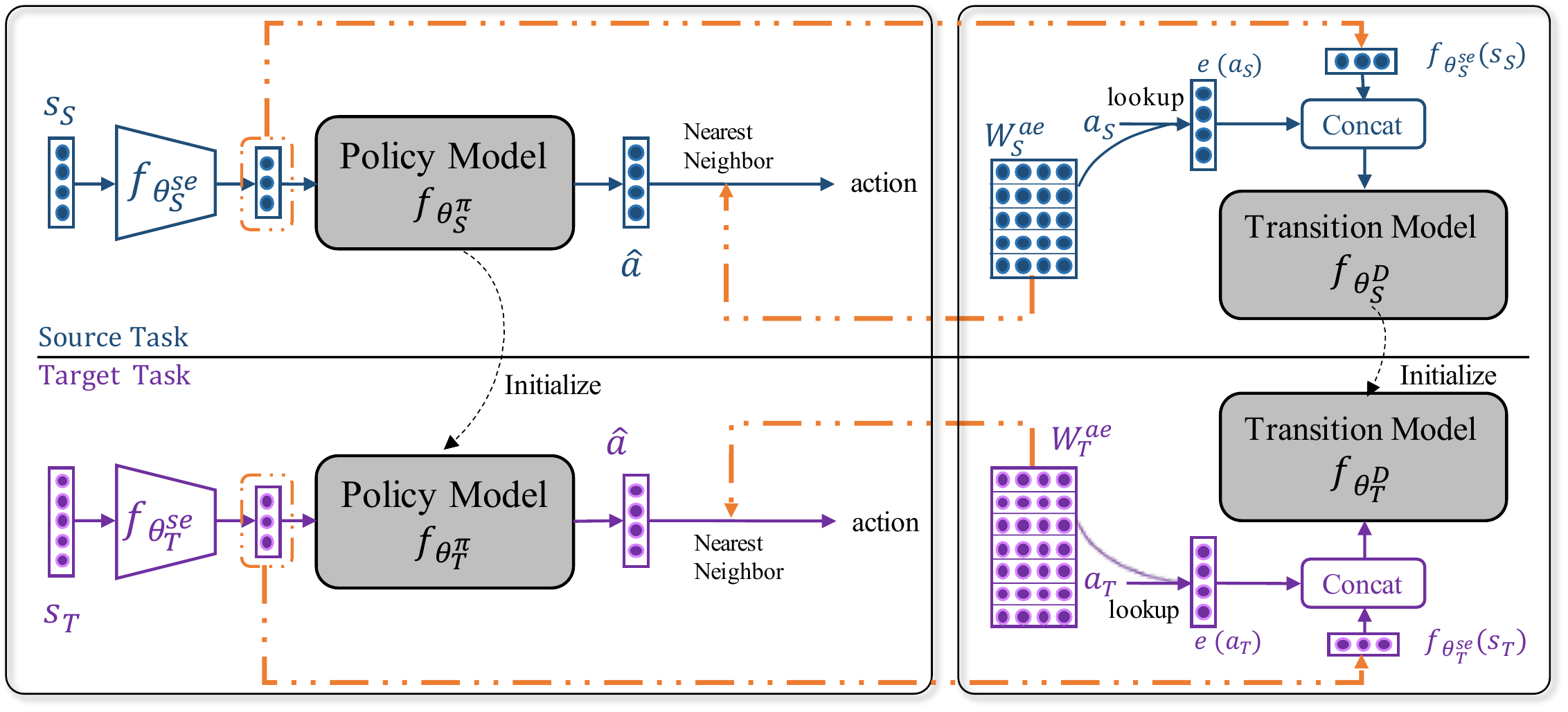}
     \caption{The architecture of TRACE for tasks with different state spaces. When transferring to the target task, the parameters in grey grids are transferred as initialization. The brown dashes mean that the data is used by the other part, but the gradient does not propagate across the module. The number of circles denotes the size of the vector, and the subscripts $S$ and $T$ denote source and target task, respectively.}
     \label{fig: arch}
     \vspace{-1em}
\end{figure*}

\begin{equation}
     \label{eq: loss}
     \begin{aligned}
          %_{\tau \sim \pi}
          \mathcal{L}(\theta^D, W^{ae}) = & \mathop{\mathbb{E}}_{s_t, a_t, s_{t + 1}} \big{[} \  ||\tilde{s}_{t+1} - s_{t+1}||_2^2
          \\ &+ \beta D_{KL} \big{(}\mathcal{N}(\mu_t,\sigma_t) \ || \ \mathcal{N}(0, I) \big{)} \big{]}
     \end{aligned}
\end{equation}
where $\theta^D =\{\theta^{D_1},\theta^{D_2}\}$, $\beta$ is a scaling factor.
%Note that the latent variable $z_t$ is not necessary for deterministic environments, and in this way, the loss function reduces to MSE loss. In our experiments, it is adopted only in combat tasks, since the other two sets of environments are deterministic. Moreover, 
Note that, if the transition of tasks is non-markovian, we can apply a recurrent transition model (such as LSTM \cite{hochreiter1997long}) to learn the dynamics more accurately.

\begin{algorithm}[!t]
     \caption{\label{alg: train} TRACE training algorithm on source task}
     \begin{algorithmic}[1]
          % \item[\textbf{Phase 1: Train policy on source task}]
          \STATE Randomly initialize the policy $f_{\theta^\pi}$, state
          embedding $f_{\theta^{se}}$, transition model $f_{\theta^D}$, and action embeddings $W^{ae}$

          /*\ \  State embedding is optional. In same-domain transfer, we set  $f_{\theta^{se}}(s) = s$  \ \  */
          \STATE Initialize replay buffer $\mathcal{B}$
          \FOR{ $episode=1$ to $L$}
          \STATE Receive initial state $s_1$ from environment
          \FOR{timestep $t=1$ to $T$}
          \STATE Select action $\hat{a}_t = f_{\theta^\pi}(f_{\theta^{se}}(s_t))$, $a_t = g(\hat{a}_t)$ according to current policy and action embeddings
          \STATE Execute action $a_t$, receive reward $r_t$, and observe new state $s_{t+1}$
          \STATE Add tuple $(s_t, \hat{a}_t, r_t, s_{t + 1})$ to $\mathcal{B}$
          \STATE Sample random batch from $B \overset{i.i.d.}{\sim} \mathcal{B}$
          \STATE Update $\theta^\pi$ and $\theta^{se}$ according to SAC training loss

          \STATE Sample random batch from $B' \overset{i.i.d.}{\sim} \mathcal{B}$ and calculate state embedding $f_{\theta^{se}}(s)$ for each state in the batch
          \STATE Update $\theta^D$ and $W^{ae}$ over Equation. (\ref{eq: loss})
          \ENDFOR
          %  \STATE Store trajectory $\{s_t, a_t\}_{t=1}^{T}$ in $\mathcal{B}$
          \ENDFOR
          \RETURN $\theta^D$, $\theta^\pi$, $\theta^{se}$, $W^{ae}$
     \end{algorithmic}
     % \begin{algorithmic}[1]
     %     \item[\textbf{Phase 2: Transfer policy to target task}]
     %     \STATE Initialize the policy $f_{\theta_T^\pi}$ and transition model $f_{\theta_T^D}$ where $\theta_T^\pi = \theta^\pi$ and $\theta_T^D = \theta^D$, and randomly initialize state embedding $f_{\theta_T^{se}}$ and action embeddings $W_T^{ae}$
     %     \STATE Train model according to line 2-14 in \textbf{Phase 1}.
     % \end{algorithmic}

\end{algorithm}

\begin{algorithm}[!t]
     \caption{\label{alg: transfer} TRACE transfer algorithm on target task}
     \textbf{Input:} Parameters $\theta_{S}^D$, $\theta_{S}^\pi$ from source task

     \begin{algorithmic}[1]
          \STATE Initialize the policy $f_{\theta^\pi}$ and transition model $f_{\theta^D}$ where $\theta^\pi = \theta_{S}^\pi$ and $\theta^D = \theta_{S}^D$, and randomly initialize state embedding $f_{\theta^{se}}$ and action embeddings $W^{ae}$

          /*\ \  In same-domain transfer, we set  $f_{\theta^{se}}(s) = s$  \ \  */
          \STATE Train model according to line 2-15 in Algorithm \ref{alg: train}
          \RETURN $\theta^D$, $\theta^\pi$, $\theta^{se}$, $W^{ae}$
     \end{algorithmic}
\end{algorithm}

\subsection{Policy Training and Transfer}

\subsubsection{Train Policy with Action Embeddings}
In this section, we describe the training process of TRACE combined with SAC. % , namely TRACE-SAC
It is noteworthy that TRACE is not limited to SAC. It can be extended to any other RL algorithms with appropriate adaptions.

Algorithm \ref{alg: train} outlines the training process on the source task of TRACE (state embedding is introduced in Section Cross-Domain Transfer).
First, we initialize the action embeddings and the network parameters of the policy model and the transition model (Line 1). During the training process, to select an action at each timestep,  the output of the policy model (continuous embedding space) should be mapped to the original discrete action space.
%one of the conventional methods is that the policy outputs over actions within a continuous space and maps the output to the original discrete action space 
In this paper, we use a standard nearest neighbor algorithm to do the mapping(Line 6) \cite{dulac2015deep}.
Specifically, the policy parameterized by $\theta^\pi$ outputs a proto-action $\hat{a} = f_{\theta^\pi}(s)$ for a given state $s$, $\hat{a} \in \mathbb{R}^d$.
Then the real action performed is chosen by a nearest neighbor in the learned action embeddings:
\begin{equation*}
     % \begin{aligned}
     g(\hat{a}) = \mathrm{argmin}_{a \in \mathcal{A}} \  ||\hat{a} - e(a)||_2
     % \end{aligned}
\end{equation*}
where $g(\cdot)$ is a mapping from a continuous space to a discrete space.
It returns an action in $\mathcal{A}$ that is closest to proto-action $\hat{a}$ in embedding space by $L_2$ distance.
The agent executes action $a_t = g(\hat{a}_t)$, receives reward $r_t$, observes next state $s_{t+1}$, and stores the transition $(s_t, \hat{a}_t, r, s_{t+1})$ to the replay buffer $\mathcal{B}$ (Lines 7-8).
Then it updates the policy model and the action embeddings and transition model accordingly following SAC loss \cite{haarnoja2018soft} and Equation (\ref{eq: loss}) (Lines 9-12).

\subsubsection{Same-Domain Transfer}
With a trained source policy, we now describe how to transfer the policy to the target task.
For better understanding, we start from a simple setting where the state spaces of source and target tasks remain the same while the action spaces are different, and we call it the same-domain transfer.
For example, a character in a game carries different sets of skills to perform the same task. Each set of skills contains different skills in terms of number and type.
Within this setting, we can directly transfer policy $f_{\theta^\pi_S}$ and fine-tune on the target task.
The agent should behave similarly when facing the same state, and the nearest-neighbor algorithm will find the most relevant skills in the target task if action embeddings of the source and target tasks are well aligned.
To achieve that, we transfer the transition model's parameters $\theta^D_S$ learned from the source task and freeze them. Then the action embeddings of the target task $W^{ae}_T \in \mathbb{R}^{|\mathcal{A}_T| \times d}$ are optimized according to Equation \ref{eq: loss}.  In this way, we can align action embeddings of the tasks, which is also validated in our experiments. %because we assume they share similar dynamics.  
%We show this in our experiments.

%Meanwhile, we can freeze the transferred transition model parameters as described before, which assures the actions are embedded into the same latent space.
%Therefore, the transferred policy can adapt to the target task quickly. 
%We will show this in our experiment part.
% When learning action embeddings of a target task, we first randomly initialize an embedding matrix $W_T^{ae} \in \mathbb{R}^{|\mathcal{A}_T| \times d}$ for the target task and optimize it according to Equation \ref{eq: loss}.
% To accelerate the training process, we can transfer the transition model's parameters $\theta_S^{D}$ learned from the source task as initialization and train or freeze the parameters on the target task. Because we assume that the source and the target tasks share similar dynamics, this can not only speed up the training of the action embeddings but also align action embeddings of the source and the target tasks. % We show this in our experiment.

\vspace{-4pt}
\subsubsection{Cross-Domain Transfer}
In cross-domain transfer, where tasks differ in both state and action spaces, the transition model can not be reused since the dimensions of states are different between source and target tasks.
The premise of reusing the transition model is that states can be embedded into the same or similar space with the same size.
%we can try to embed the states of tasks into a common feature space and train the transition model based on extracted features.   
Thus, the input of the transition model becomes a tuple $(f_{\theta^{se}}(s_t), a_t, f_{\theta^{se}}(s_{t + 1}))$, where $f_{\theta^{se}}(\cdot)$ denotes a non-linear function with parameter $\theta^{se}$, mapping the original state space into a common space, called state embedding.
In this work, we train the state embedding along with the policy.
Note that the two modules (policy model and transition model) become interdependent --- transition model needs state embeddings as training input, and policy requires action embeddings to select actions.
Therefore, we train two modules together, which can also increase data utilization.
The architecture is shown in Figure \ref{fig: arch}
% Overall, we outline the training process on the source task in in Algorithm \ref{alg: train}, and Figure \ref{fig: arch} shows the architecture. 
% As seen, for the target task, we reinitialize the state embedding $f_{\theta_T^{se}}$ and action embeddings $W_T^{ae}$, while the policy $f_{\theta_T^\pi}$ and the transition model $f_{\theta_T^D}$ are initialized with $\theta_S^\pi$ and $\theta_S^D$.

When training on the target task, we initialize the RL policy $f_{\theta_T^\pi}$ and the transition model $f_{\theta_T^D}$ with parameters $\theta_S^\pi$ and $\theta_S^D$ from the source task. At the same time, the state embedding $f_{\theta_T^{se}}$ and action embeddings $W_T^{ae}$ are randomly reinitialized.
%ignoring the state embedding $\phi_S$ that is trained to encode the source task efficiently.
Then we train the network as on the source task (Lines 2-15 in Algorithm \ref{alg: train}).
Algorithm \ref{alg: transfer} outlines the transfer process.
Unlike the same-domain transfer, the transition model parameters are not frozen but finetuned. This is because the inconsistency in the transition model increases when state dimensions are different, which leads to unstable training. We also verify this in our experiments shown in the appendix.%and the input of state embeddings is trained over time. Additionally, according to our preliminary experiment, it leads to unstable training by freezing the parameters.
% Besides, the state embedding of the target task is trained together with the RL policy.
% This might be seen as a kind of adaptation where the state embedding $f_{\theta_T^{se}}$ is regularized by policy and transition model. 
%While there exist some promising methods, such as adversarial autoencoder \cite{carr2019domain}, to align the state representation function $f_{\theta_T^{se}}$ with $f_{\theta_S^{se}}$ in the pre-training process, it's not the main point in this paper.

\begin{figure}
     \centering
     \includegraphics[width=240pt]{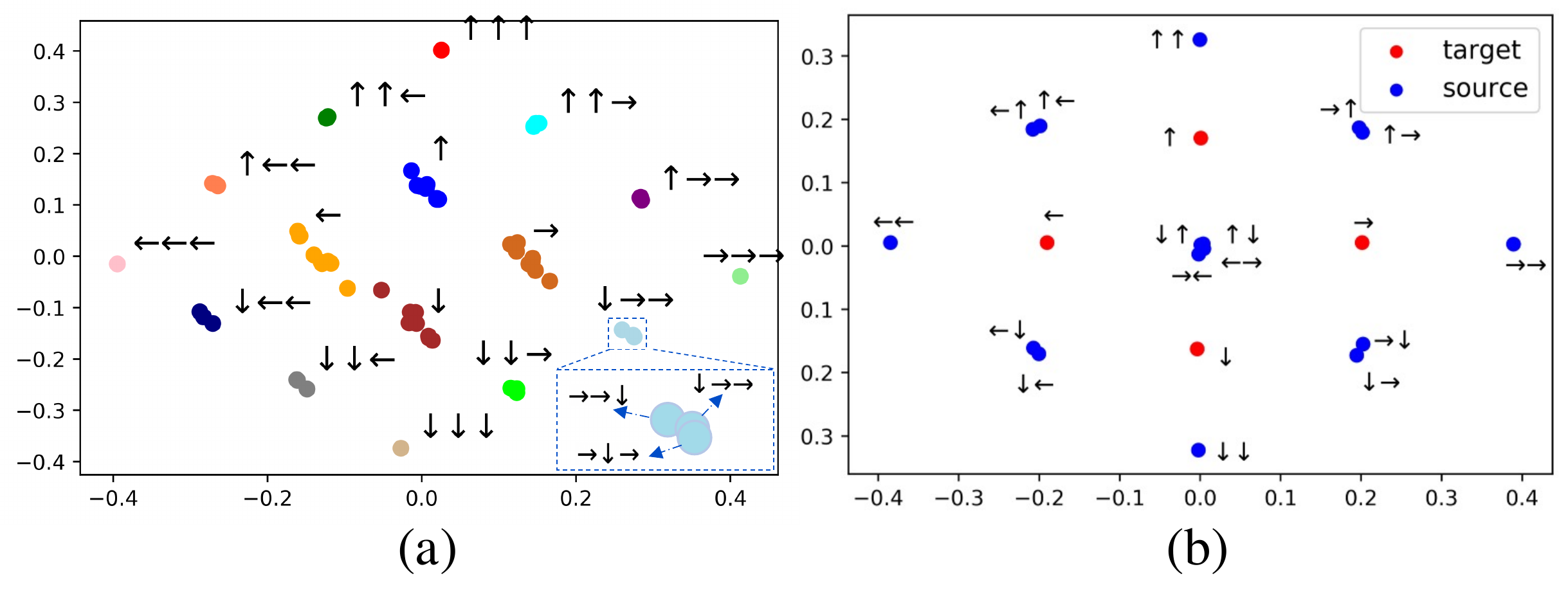}
     \vspace{-10pt}
     \caption{The learned embeddings projected into 2D space via PCA. Each dot in (a) represents an action embedding in gridworld. We show the action effect of each group.  (b) Action embeddings of the source task (2-step gridworld) and the target task (1-step gridworld).}
     \label{fig: embed}
     \vspace{-10pt}
\end{figure}

\begin{figure*}
     \centering
     \includegraphics[height=15pt]{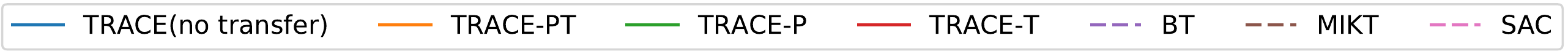}

     \vspace{-0.7em}
     \subfigure[task $n=1$ (source task $n=3$)]{ % 
          \label{fig: gridworld-result:a}
          \includegraphics[width=0.25\textwidth]{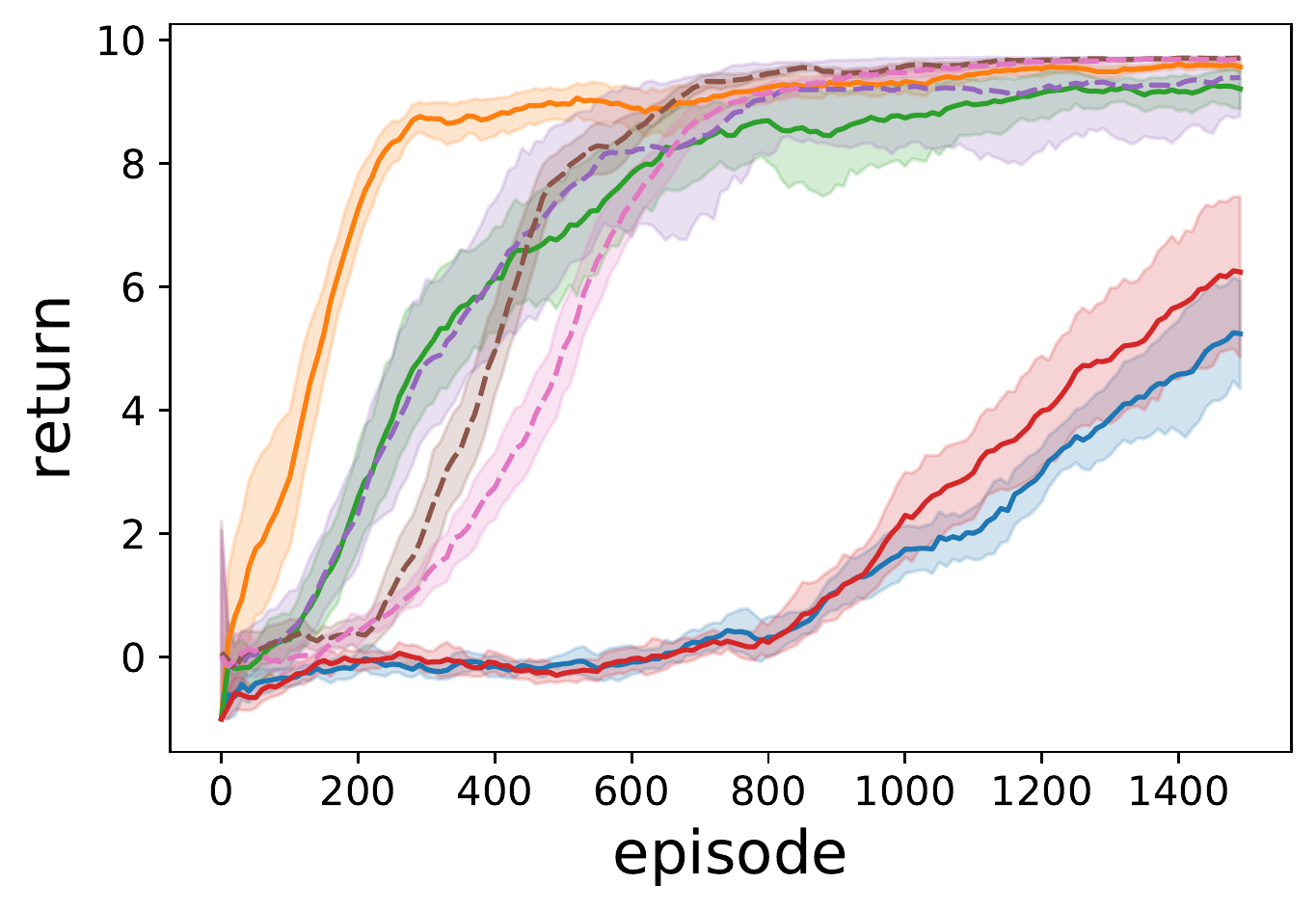}
     }
     \subfigure[task $n=2$ (source task $n=1$)]{
          \label{fig: gridworld-result:b}
          \includegraphics[width=0.25\textwidth]{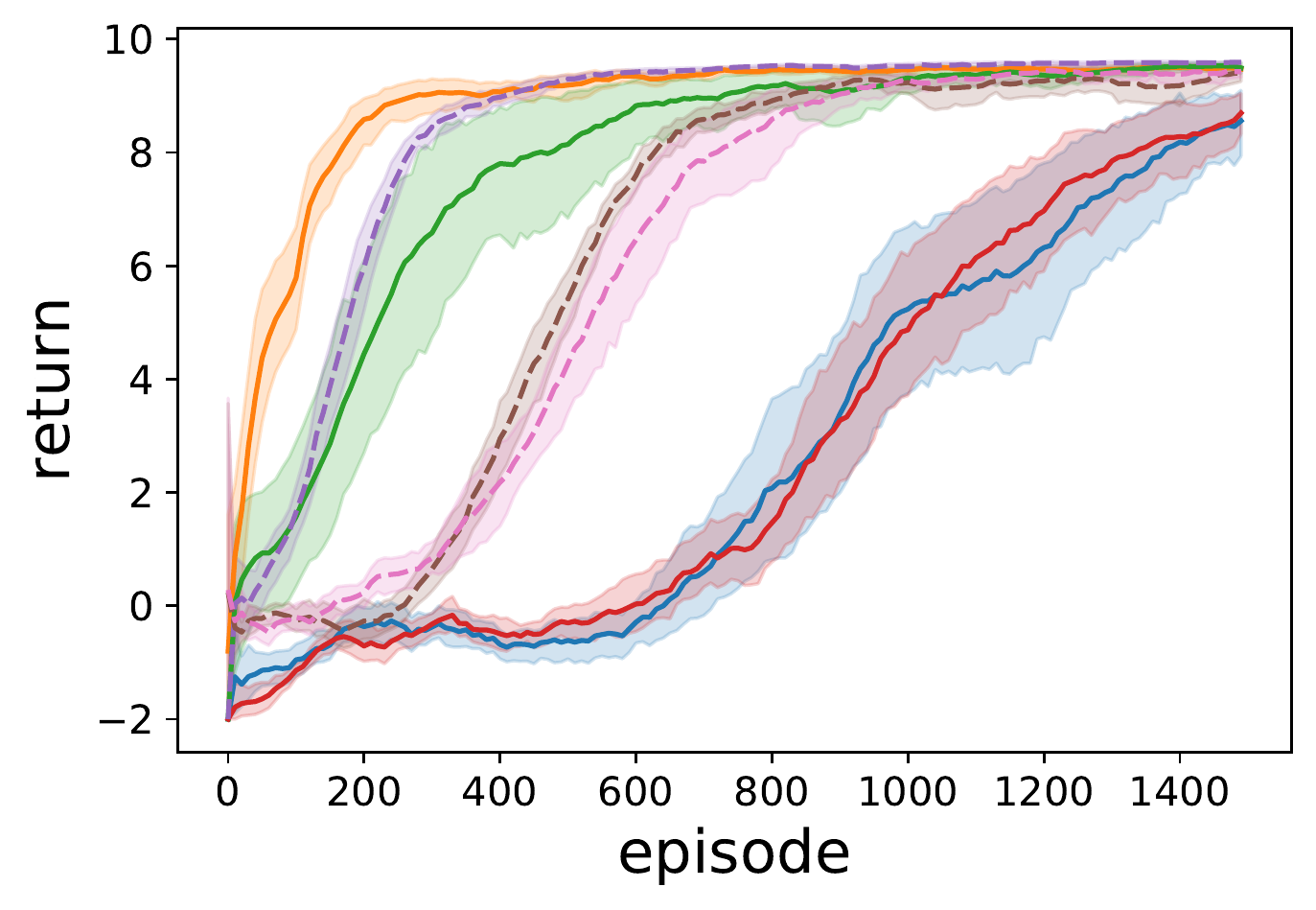}
     }
     \subfigure[task $n=3$ (source task $n=2$)]{
          \label{fig: gridworld-result:c}
          \includegraphics[width=0.25\textwidth]{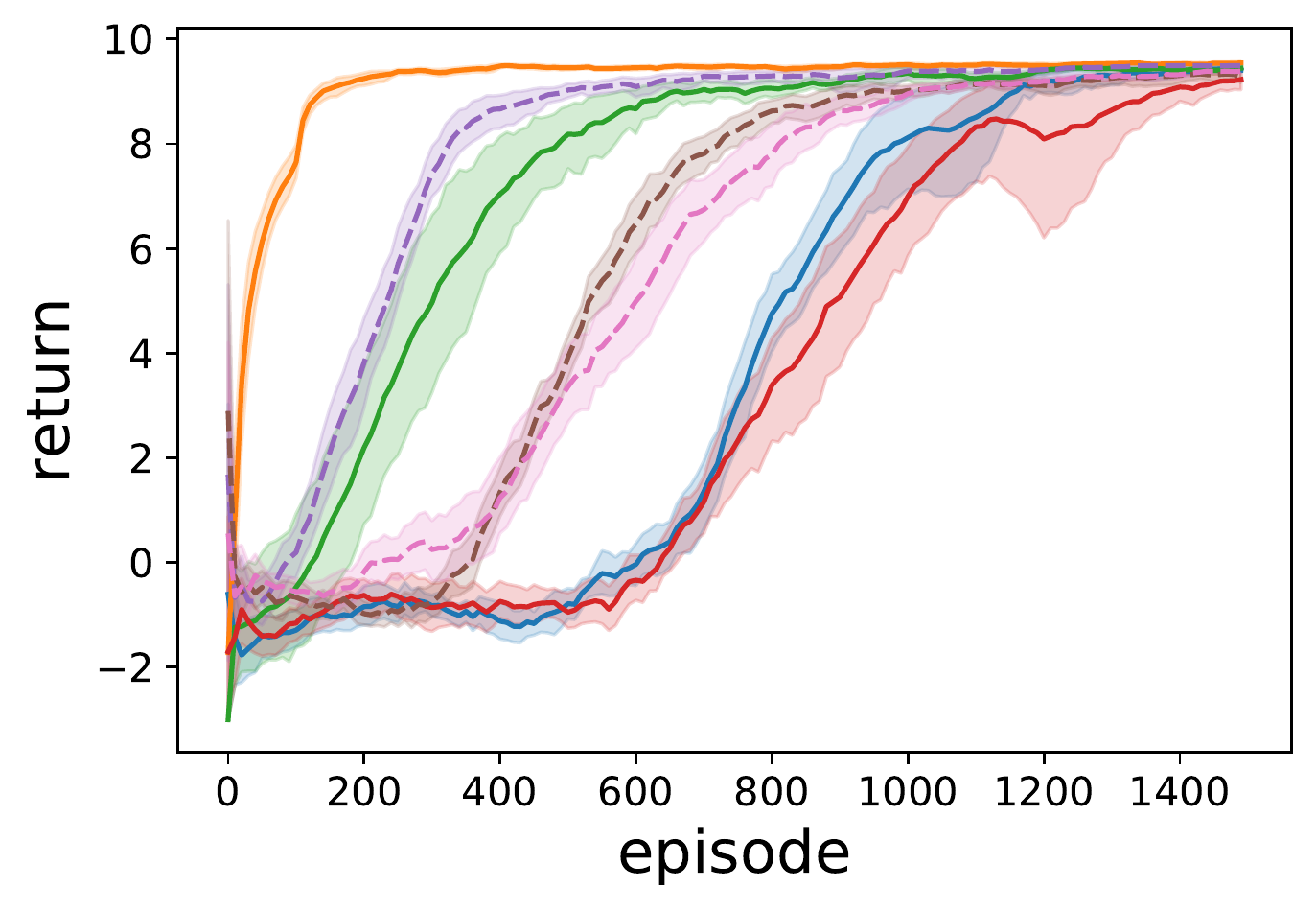}
     }
     \vspace{-10pt}
     \caption{Experiment results on gridworld environments. The solid lines denote our method, and the dashed lines represent the compared methods. The shaded areas are bootstrapped 95\% confidence intervals.}
     \label{fig: grid-result}
     \vspace{-10pt}
\end{figure*}

\section{Experiments}
In this section, we empirically investigate the feasibility of learning action embeddings with the transition model and assess the effectiveness of TRACE.
% We use three sets of sequential decision-making tasks to evaluate our methods: n-step gridworld navigation tasks, discretized continuous control tasks, and combat tasks in a commercial game.

We compare our method, abbreviated as \textit{TRACE-PT}, which means to transfer both the Policy model (P) and the Transition model (T), with three baselines:
a) \textit{SAC} \cite{haarnoja2018soft}, which learns from scratch on target tasks;
b) A basic transfer (\textit{BT}) strategy, which replaces the input and output layers of the neural network learned from the source task with new learnable layers that match the dimensions required of the target task and fine-tunes the whole network.
c) \textit{MIKT} \cite{wan2020mutual}, which leverages policies learned from source tasks as teachers to enhance the learning of target tasks;
Note that \textit{BT} can see as an ablation of our method that does not use action embedding.
We also report the result of our method that learns from scratch on the target task, denoted by \textit{TRACE(no transfer)}.
All the methods are based on SAC, and the results are averaged over 10 individual runs.

Appendix A provides detailed descriptions of environments and hyperparameters used in the experiments, and more experimental results are available in Appendix B.

\subsection{N-Step Gridworld}
We first validate our methods in an $11 \times 11$ gridworld, in which an agent needs to reach a randomly assigned goal position.
The agent could perform 4 atomic actions: \emph{Up}, \emph{Down}, \emph{Left}, and \emph{Right} at each step.
Once we consider combo moves in successive $n$-step, the number of actions becomes $4^n$.
We conduct experiments on three settings with $n \in \{1, 2, 3\}$.
The sizes of action spaces are 4, 16, 64, respectively.
The state of the tasks consists of the current position $(x, y)$ and the goal position $(\dot{x}, \dot{y})$.
The agent receives a -0.05 for each step and a +10 reward when the agent reaches the goal.
% The performance of the compare d methods is measured by the average return of 100 episodes.

To investigate whether action embedding can capture the semantics of actions, we set $n = 3$, the dimension of action embedding $d = 4$, and sample 10,000 transition data based on a random policy.
Figure \ref{fig: embed}(a) shows the Principal Component Analysis (PCA) projections of the resulting embeddings.
In the figure, different colors denote different action effects.
For example, there are 9 blue dots (\textcolor{blue}{$\uparrow$}), including $\uparrow\uparrow\downarrow$, $\leftarrow\uparrow\rightarrow$, etc. % $\uparrow\leftarrow\rightarrow$,
We see that the actions with the same effect are positioned closely and clustered into 16 separate groups.
What's more, those clusters show near-perfect symmetry along with the four directions in the gridworld, which means our method effectively captures the semantics of actions.
In word embeddings, the relationship between words is often discussed, such as \textit{Paris - France + Italy = Rome} \cite{mikolov2013efficient}.
In action embeddings, we can also get the similar property, such as $e(\uparrow \uparrow \leftarrow) + e(\uparrow \leftarrow \rightarrow) - e(\leftarrow \rightarrow \leftarrow) \approx e(\uparrow \uparrow \uparrow)$.

Further, we evaluate the policy transfer performance on tasks $n = \{1, 2, 3\}$.
Note that state spaces of these tasks are the same. So we freeze the parameters $\theta^D$ of the transferred transition model as described in Section 4.3.
As seen in Figure \ref{fig: grid-result},  the speed of training on target tasks with TRACE-PT outperforms that of all the other methods in all tasks.
MIKT learns faster than SAC, but slower than BT.
This is because that reusing previous parameters is more efficient than distilling regarding the same-domain transfer.
Besides, we find that SAC performs better than TRACE in all tasks because we map discrete actions into a continuous space, which makes it challenging to learn the policy, especially when the number of actions is small. Note that there are no jump-starts on those curves due to the action embeddings of target tasks are randomly initialized. However, the action embeddings adapt quickly, which results in a fast transfer.
Moreover, the action embeddings of the target task should align with the source task so that policy could have a promising performance on the target task.
To verify this, we project the embeddings of both source task $n=2$ and target task $n=1$ into 2D space. As shown in Figure \ref{fig: embed}(b), the embeddings of the source task and the target task are well aligned, and it is even observed that $e(\textcolor{red}{\uparrow}) \approx 0.5* e(\textcolor{blue}{\uparrow \uparrow}) + 0.5* e(\textcolor{blue}{\uparrow \downarrow})$.

\begin{figure*}
     \centering
     \includegraphics[height=15pt]{fig/new_legend2.pdf}

     \vspace{-0.5em}
     \subfigure[mP (source task mDP)]{
          \label{fig: mujoco-result:a}
          \includegraphics[width=0.25\textwidth]{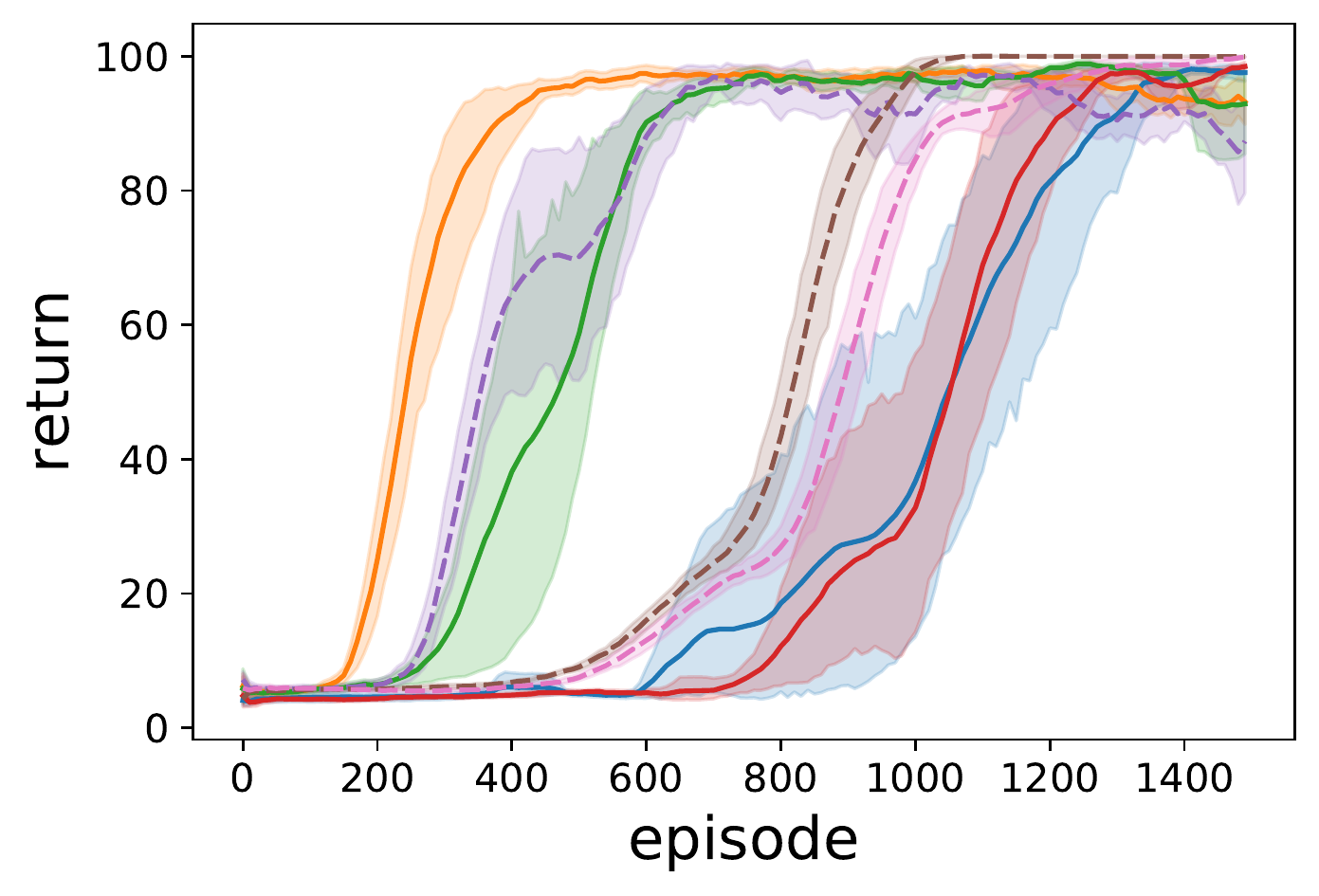}
     } \hspace{-1.15em}
     \subfigure[rP (source task rDP)]{
          \label{fig: mujoco-result:b}
          \includegraphics[width=0.25\textwidth]{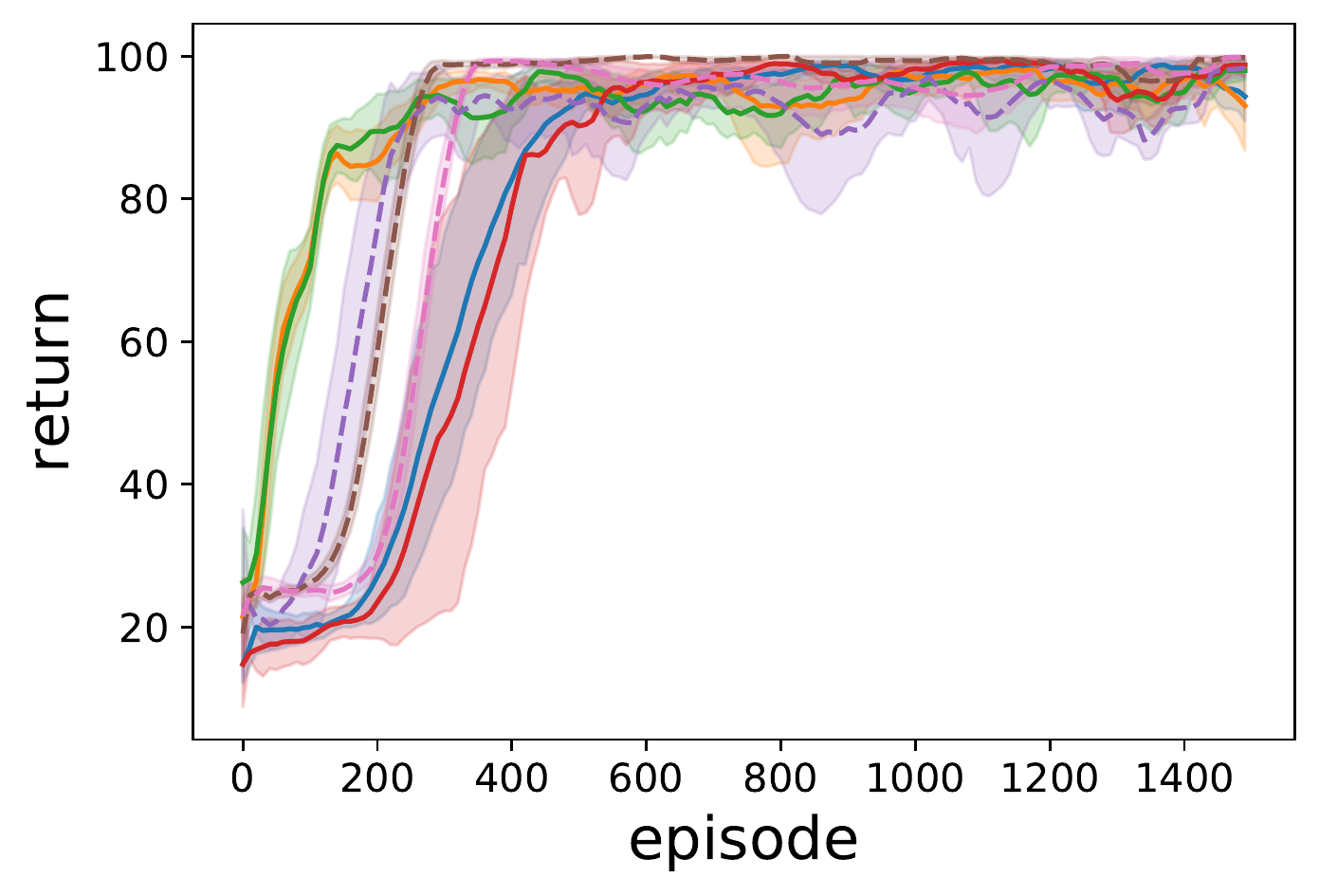}
     } \hspace{-1.15em}
     \subfigure[mDP (source task rP)]{
          \label{fig: mujoco-result:c}
          \includegraphics[width=0.25\textwidth]{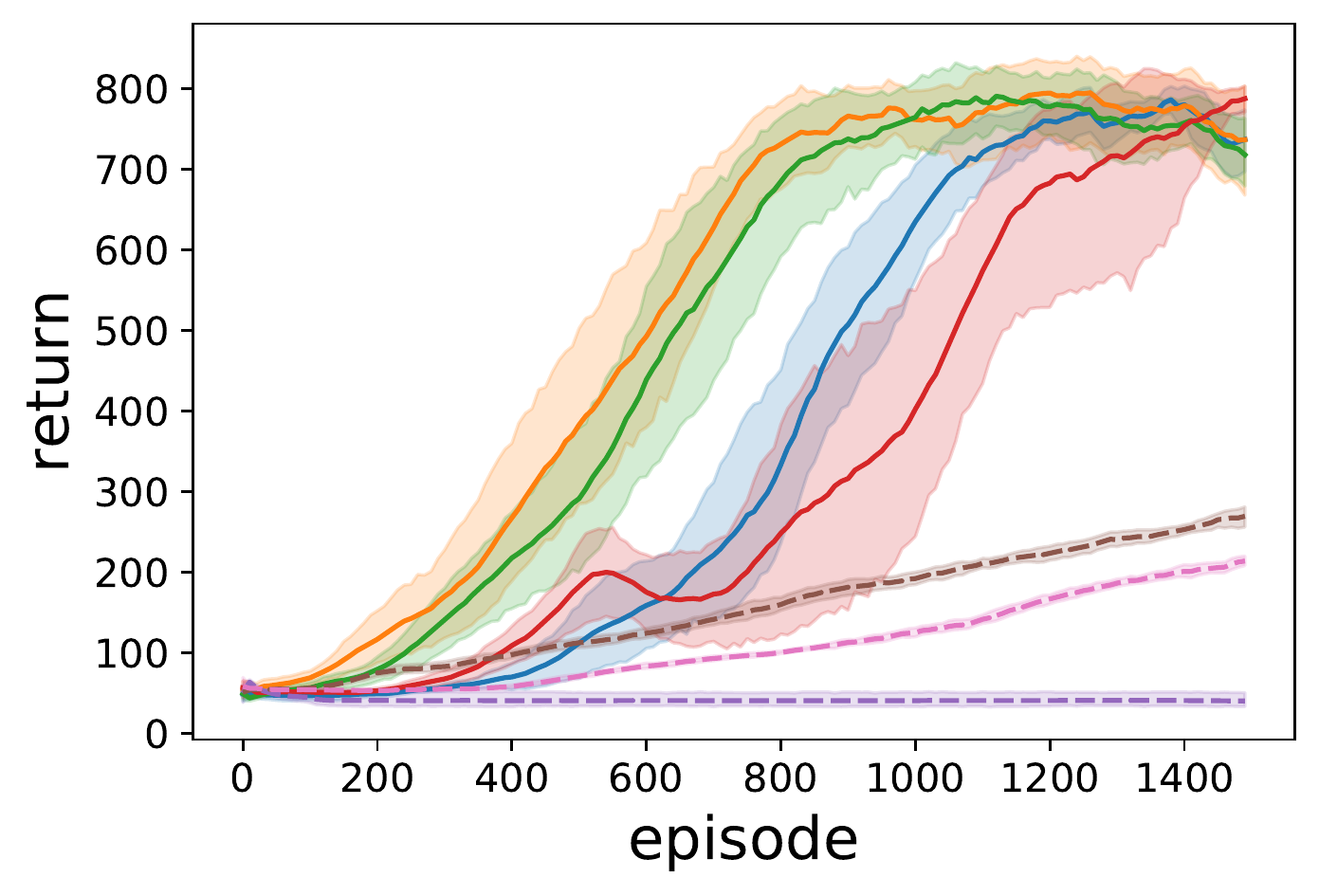}
     } \hspace{-1.15em}
     \subfigure[rDP (source task mP)]{
          \label{fig: mujoco-result:d}
          \includegraphics[width=0.25\textwidth]{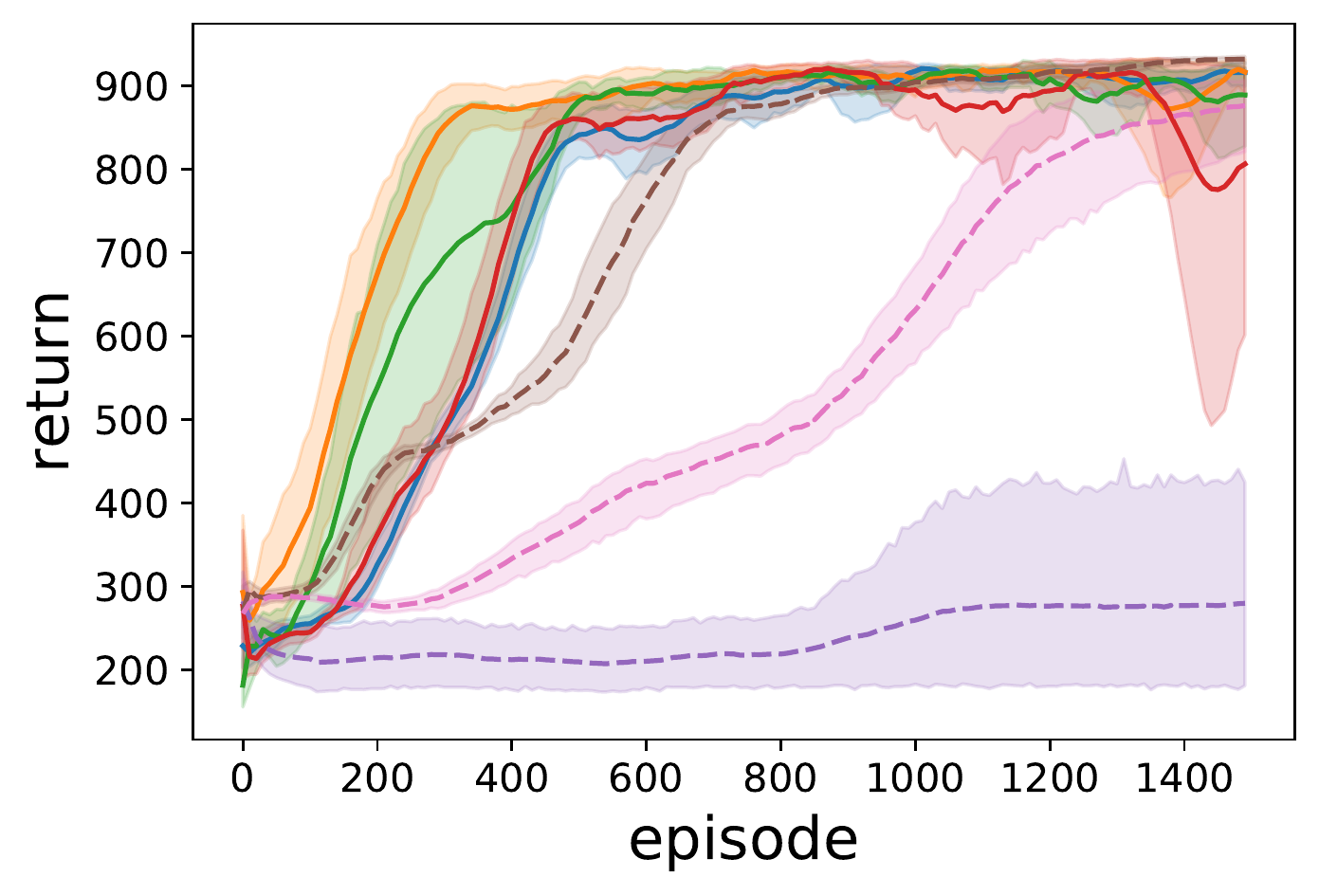}
     }
     \vspace{-10pt}
     \caption{Experiment results on Mujoco and Roboschool tasks. The solid lines denote our method, and the dashed lines represent the compared methods. The shaded areas are bootstrapped 95\% confidence intervals.}
     \label{fig: mujoco-result}
     \vspace{-10pt}
\end{figure*}

\subsection{Mujoco and Roboschool}
Next, we consider a more difficult cross-domain transfer setting, in which state spaces are different as well as action spaces.
We conduct experiments among four environments, \textit{InvertedPendulum} and \textit{InvertedDoublePendulum} in \textbf{Mujoco} \cite{todorov2012mujoco} and \textbf{Roboschool}, respectively, denoted by \textit{mP} (mujoco Pendulum), \textit{mDP} (mujoco Double Pendulum), \textit{rP} (roboschool Pendulum) and \textit{rDP} (roboschool Double Pendulum) for short.
Currently, our methods are only suitable for discrete action spaces.
So we discretize the original $m$-dimension continuous control action space into $k$ equally spaced values on each dimension, resulting in a discrete action space with $|\mathcal{A}| = k^m$ actions.
The detailed configurations and descriptions of the environments are summarized in Appendix A.3 , and the learned action embeddings of the environments are shown in Figure 4 of Appendix.
% , due to the limit of pages.

In this experiment, our method is evaluated on four transfer tasks. We first try to transfer policy from \textit{mDP} to \textit{mP} and \textit{rDP} to \textit{rP}, where the source and the target tasks are still in the same underlying physical engine, and the target tasks are easier than the source ones. Further, we transfer policy from \textit{mP} to \textit{rDP} and \textit{rP} to \textit{mDP}, the source and the target tasks are in different physical engines, and the target tasks are more challenging than the source tasks.
Figure \ref{fig: mujoco-result} depicts the results of cross-domain transfer.
Overall, TRACE-PT still learns faster than SAC and BT in all tasks, and MIKT performs slightly better than SAC.
We can see that in Figure \ref{fig: mujoco-result:a} and \ref{fig: mujoco-result:b}, though BT has a promising performance, it fails to learn the target task in more challenging transfer tasks and even has negative transfer, shown in Figure \ref{fig: mujoco-result:c} and \ref{fig: mujoco-result:d}.
According to our extended experiments shown in Figure 3 of Appendix B, BT only performs well in the simplest cases. In most cases, it leads to a negative transfer. In contrast, TRACE-PT can handle all the transfer tasks and accelerate learning. This indicates that our approach is superior to BT and the advantages are more apparent when in more challenging transfer tasks.

\subsection{Combat Tasks in a Commercial Game}
To inspect our methods' potential in more practical problems, we validate them on a one-versus-one combat scenario in a commercial game where the agent can carry different sets of skills to fight against a build-in opponent.
In this experiment, we select two classes named \emph{She Shou} (SS) and \emph{Fang Shi} (FS). % , and both the two classes are DPS
The state representations are extracted manually and consist of information about the controlled agent and build-in opponent, forming two vectors with 48 and 60-dimensional, respectively. The sizes of action spaces are 10 for both classes, containing their unique skills and standard operations, such as move and attack.
The agent receives positive rewards for damaging and winning, and negative for taking damage and losing. Besides, the agent is punished for choosing unready skills.

We first verify whether our method can learn reasonable action embeddings in complex games.
\begin{wrapfigure}{r}{0.23\textwidth}
     \centering
     \includegraphics[width=0.25\textwidth]{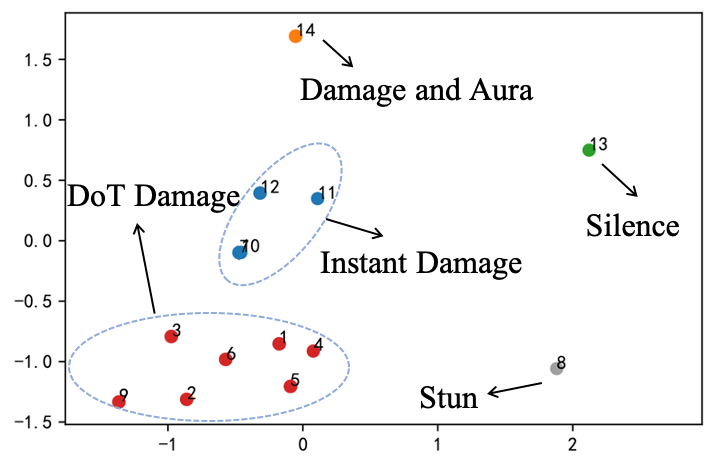}
     \caption{PCA projections of learned action embeddings. Each dot represents an action embedding of skill in the game.}
     \vspace{-10pt}
     \label{fig: qnyh_embed}
\end{wrapfigure}
we randomly sample 5 out of 15 skills and collect transition data. Table 4 of Appendix A lists the skill descriptions.
We sample 50,000 transition data to train action embeddings with $d=6$.
Figure \ref{fig: qnyh_embed} plots the result, and we notice that the skills with special effects, such as \textbf{Silence} and \textbf{Stun}, are distinguished clearly, and that damage skills are also closer to each other.
As annotated in the figure, the \textbf{DoT Damage} and \textbf{Instant Damage} are recognized as well.

\begin{figure}
     \centering
     \includegraphics[width=240pt]{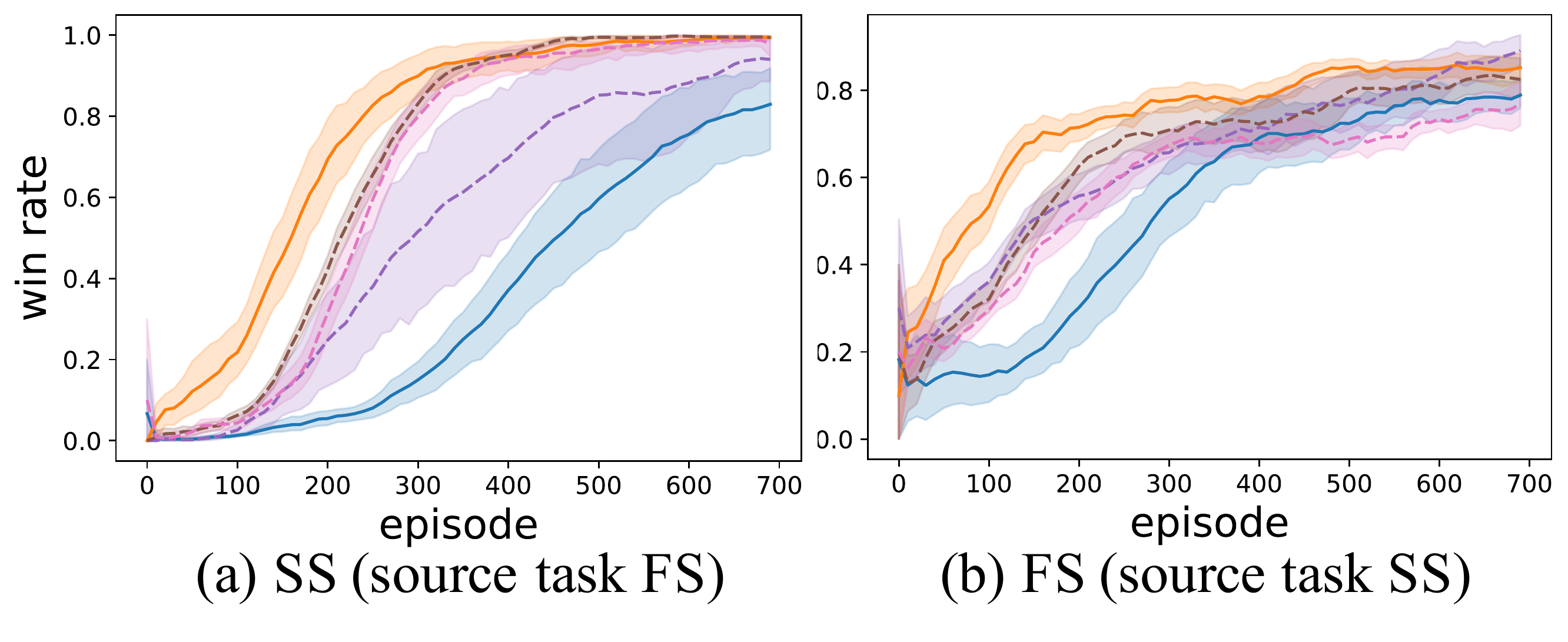}
     %  \caption{The learning curves of methods on the fighting video game. }
     %  \label{fig: qnyh-result}
     %   \subfigure[]{
     %       \label{fig: qnyh-result:a}
     %       \includegraphics[width=115pt]{fig/qnyh-sheshou-new.pdf}
     %   } \hspace{-1em}
     %   \subfigure[]{
     %     \label{fig: qnyh-result:b}
     %     \includegraphics[width=115pt]{fig/qnyh-fangshi-new.pdf}
     %   }
     \vspace{-10pt}
     \caption{Experiment results on combat tasks.}
     \label{fig: qnyh-result}
     \vspace{-1em}
\end{figure}

For policy transfer, we first train policy individually for two roles and then transfer to each other.
The performance is measured by the average winning rate of the recent 100 episodes, as plotted in Figure \ref{fig: qnyh-result}.
For clarity, we show ablation results in Appendix B.
We can see that TRACE-PT achieves a better sample efficiency compared with MIKT, while BT may leads to the negative transfer.
%We can see that \emph{TRACE-PT} and \emph{MIKT} achieves competitive sample efficiency in the tasks, while \emph{BT} increases quickly at the beginning of learning but gets slow soon and appears to be unstable. 
It proves that our method can be applied to more practical problems.
% In contrast, \emph{TRACE-PT} achieves higher sample efficiency. 

\vspace{-5pt}
\subsection{Ablations}
To better understand our method, we analyze the contribution of the transition model and the policy model to performance promotion. The ablation experiments are designed as follows:
\begin{itemize}
     \item \textit{TRACE-P}: Transfer policy model only, and randomly initialize transition model.
     \item \textit{TRACE-T}: Transfer transition model only, and train policy from scratch.
     \item \textit{BT}: Transfer policy, and do not use action embedding.
\end{itemize}

For same-domain transfer (Figure \ref{fig: grid-result}), we find that TRACE-P can also accelerate learning significantly. However, there still lies a gap between TRACE-PT and TRACE-P because it needs to learn the transition model and action embeddings anew, making the action embeddings of the target task may not align with the source task, and the policy requires more time to adapt. However, in the cross-domain transfer (Figure \ref{fig: mujoco-result}), TRACE-P achieves a competitive performance to TRACE-PT , especially in Figure \ref{fig: mujoco-result:b}.
It is easy to understand that the transition model and the action embeddings are learned from state embeddings, which are retrained on the target task. It limits the performance of transfer.
Besides, in both same-domain and cross-domain transfers, TRACE-T results in a similar performance to TRACE, which may indicate that the cost of training mainly lies in policy training and action representations can boost policy generalization.
%learning based on action embeddings is more difficult, but leads to better generalization. 
Comparing TRACE-PT and BT, we find that the proposed action embeddings can indeed facilitate policy transfer.

\vspace{-7pt}
\section{Conclusion}
In this paper, we study how to leverage action embeddings to transfer across tasks with different action spaces and/or state spaces. We propose a method to effectively learn meaningful action embeddings by training a transition model. Further, we train RL policies with action embeddings by using the nearest neighbor in the embedding space. The policy and transition model are transferred to the target task, leading to a quick adaptation of policy. Extensive experiments demonstrate that it significantly improves sample efficiency, even with different state spaces and action spaces. In the future, we will try to extend our method to continuous action spaces and align the state embeddings with additional restrictions.

% \clearpage
\small
\bibliographystyle{named}
\bibliography{ijcai21}

\begin{thebibliography}{}

\bibitem[\protect\citeauthoryear{Ammar \bgroup \em et al.\egroup
  }{2015}]{ammar2015unsupervised}
Haitham~Bou Ammar, Eric Eaton, Paul Ruvolo, and Matthew~E Taylor.
\newblock Unsupervised cross-domain transfer in policy gradient reinforcement
  learning via manifold alignment.
\newblock In {\em Twenty-Ninth AAAI Conference on Artificial Intelligence},
  2015.

\bibitem[\protect\citeauthoryear{Barreto \bgroup \em et al.\egroup
  }{2019}]{barreto2019transfer}
Andr{\'e} Barreto, Diana Borsa, John Quan, Tom Schaul, David Silver, Matteo
  Hessel, Daniel Mankowitz, Augustin {\v{Z}}{\'\i}dek, and Remi Munos.
\newblock Transfer in deep reinforcement learning using successor features and
  generalised policy improvement.
\newblock {\em arXiv preprint arXiv:1901.10964}, 2019.

\bibitem[\protect\citeauthoryear{Carr \bgroup \em et al.\egroup
  }{2019}]{carr2019domain}
Thomas Carr, Maria Chli, and George Vogiatzis.
\newblock Domain adaptation for reinforcement learning on the atari.
\newblock In {\em Proceedings of the 18th International Conference on
  Autonomous Agents and MultiAgent Systems}, pages 1859--1861, 2019.

\bibitem[\protect\citeauthoryear{Chandak \bgroup \em et al.\egroup
  }{2019}]{ChandakTKJT19}
Yash Chandak, Georgios Theocharous, James Kostas, Scott~M. Jordan, and
  Philip~S. Thomas.
\newblock Learning action representations for reinforcement learning.
\newblock In {\em ICML}, pages 941--950, 2019.

\bibitem[\protect\citeauthoryear{Dulac-Arnold \bgroup \em et al.\egroup
  }{2015}]{dulac2015deep}
Gabriel Dulac-Arnold, Richard Evans, Hado van Hasselt, Peter Sunehag, Timothy
  Lillicrap, Jonathan Hunt, Timothy Mann, Theophane Weber, Thomas Degris, and
  Ben Coppin.
\newblock Deep reinforcement learning in large discrete action spaces.
\newblock {\em arXiv preprint arXiv:1512.07679}, 2015.

\bibitem[\protect\citeauthoryear{Finn \bgroup \em et al.\egroup
  }{2017}]{finn2017model}
Chelsea Finn, Pieter Abbeel, and Sergey Levine.
\newblock Model-agnostic meta-learning for fast adaptation of deep networks.
\newblock In {\em ICML}, pages 1126--1135, 2017.

\bibitem[\protect\citeauthoryear{Goyal \bgroup \em et al.\egroup
  }{2017}]{goyal2017z}
Anirudh Goyal Alias~Parth Goyal, Alessandro Sordoni, Marc-Alexandre
  C{\^o}t{\'e}, Nan~Rosemary Ke, and Yoshua Bengio.
\newblock Z-forcing: Training stochastic recurrent networks.
\newblock In {\em Advances in neural information processing systems}, pages
  6713--6723, 2017.

\bibitem[\protect\citeauthoryear{Gupta \bgroup \em et al.\egroup
  }{2017}]{gupta2017learning}
Abhishek Gupta, Coline Devin, YuXuan Liu, Pieter Abbeel, and Sergey Levine.
\newblock Learning invariant feature spaces to transfer skills with
  reinforcement learning.
\newblock {\em arXiv preprint arXiv:1703.02949}, 2017.

\bibitem[\protect\citeauthoryear{Haarnoja \bgroup \em et al.\egroup
  }{2018}]{haarnoja2018soft}
Tuomas Haarnoja, Aurick Zhou, Pieter Abbeel, and Sergey Levine.
\newblock Soft actor-critic: Off-policy maximum entropy deep reinforcement
  learning with a stochastic actor.
\newblock {\em arXiv preprint arXiv:1801.01290}, 2018.

\bibitem[\protect\citeauthoryear{Hochreiter and
  Schmidhuber}{1997}]{hochreiter1997long}
Sepp Hochreiter and J{\"u}rgen Schmidhuber.
\newblock Long short-term memory.
\newblock {\em Neural computation}, 9(8):1735--1780, 1997.

\bibitem[\protect\citeauthoryear{Jain \bgroup \em et al.\egroup
  }{2020}]{jain2020generalization}
Ayush Jain, Andrew Szot, and Joseph Lim.
\newblock Generalization to new actions in reinforcement learning.
\newblock In {\em International Conference on Machine Learning}, pages
  4661--4672. PMLR, 2020.

\bibitem[\protect\citeauthoryear{Kingma and Welling}{2013}]{kingma2013auto}
Diederik~P Kingma and Max Welling.
\newblock Auto-encoding variational bayes.
\newblock {\em arXiv preprint arXiv:1312.6114}, 2013.

\bibitem[\protect\citeauthoryear{Levine \bgroup \em et al.\egroup
  }{2016}]{levine2016end}
Sergey Levine, Chelsea Finn, Trevor Darrell, and Pieter Abbeel.
\newblock End-to-end training of deep visuomotor policies.
\newblock {\em JMLR}, 17(1):1334--1373, 2016.

\bibitem[\protect\citeauthoryear{Liu \bgroup \em et al.\egroup
  }{2019}]{liu2019knowledge}
Iou~Jen Liu, Jian Peng, and Alexander~G Schwing.
\newblock Knowledge flow: Improve upon your teachers.
\newblock In {\em 7th International Conference on Learning Representations,
  ICLR 2019}, 2019.

\bibitem[\protect\citeauthoryear{Ma \bgroup \em et al.\egroup
  }{2018}]{ma2018universal}
Chen Ma, Junfeng Wen, and Yoshua Bengio.
\newblock Universal successor representations for transfer reinforcement
  learning.
\newblock {\em arXiv preprint arXiv:1804.03758}, 2018.

\bibitem[\protect\citeauthoryear{Mikolov \bgroup \em et al.\egroup
  }{2013}]{mikolov2013efficient}
Tomas Mikolov, Kai Chen, Greg Corrado, and Jeffrey Dean.
\newblock Efficient estimation of word representations in vector space.
\newblock {\em arXiv preprint arXiv:1301.3781}, 2013.

\bibitem[\protect\citeauthoryear{Mnih \bgroup \em et al.\egroup
  }{2015}]{mnih2015human}
Volodymyr Mnih, Koray Kavukcuoglu, David Silver, Andrei~A Rusu, Joel Veness,
  Marc~G Bellemare, Alex Graves, Martin Riedmiller, Andreas~K Fidjeland, Georg
  Ostrovski, et~al.
\newblock Human-level control through deep reinforcement learning.
\newblock {\em Nature}, 518(7540):529, 2015.

\bibitem[\protect\citeauthoryear{Raiman \bgroup \em et al.\egroup
  }{2019}]{raiman2019neural}
Jonathan Raiman, Susan Zhang, and Christy Dennison.
\newblock Neural network surgery with sets.
\newblock {\em arXiv preprint arXiv:1912.06719}, 2019.

\bibitem[\protect\citeauthoryear{Silver \bgroup \em et al.\egroup
  }{2016}]{silver2016mastering}
David Silver, Aja Huang, Chris~J Maddison, Arthur Guez, Laurent Sifre, George
  Van Den~Driessche, Julian Schrittwieser, Ioannis Antonoglou, Veda
  Panneershelvam, Marc Lanctot, et~al.
\newblock Mastering the game of go with deep neural networks and tree search.
\newblock {\em nature}, 529(7587):484, 2016.

\bibitem[\protect\citeauthoryear{Taylor and Stone}{2009}]{taylor2009transfer}
Matthew~E Taylor and Peter Stone.
\newblock Transfer learning for reinforcement learning domains: A survey.
\newblock {\em JMLR}, 10(Jul):1633--1685, 2009.

\bibitem[\protect\citeauthoryear{Taylor \bgroup \em et al.\egroup
  }{2007}]{taylor2007transfer}
Matthew~E Taylor, Peter Stone, and Yaxin Liu.
\newblock Transfer learning via inter-task mappings for temporal difference
  learning.
\newblock {\em JMLR}, 8(Sep):2125--2167, 2007.

\bibitem[\protect\citeauthoryear{Teh \bgroup \em et al.\egroup
  }{2017}]{teh2017distral}
Yee Teh, Victor Bapst, Wojciech~M Czarnecki, John Quan, James Kirkpatrick, Raia
  Hadsell, Nicolas Heess, and Razvan Pascanu.
\newblock Distral: Robust multitask reinforcement learning.
\newblock In {\em Advances in Neural Information Processing Systems}, pages
  4496--4506, 2017.

\bibitem[\protect\citeauthoryear{Tennenholtz and Mannor}{2019}]{TennenholtzM19}
Guy Tennenholtz and Shie Mannor.
\newblock The natural language of actions.
\newblock In {\em International Conference on Machine Learning}, pages
  6196--6205, 2019.

\bibitem[\protect\citeauthoryear{Todorov \bgroup \em et al.\egroup
  }{2012}]{todorov2012mujoco}
Emanuel Todorov, Tom Erez, and Yuval Tassa.
\newblock Mujoco: A physics engine for model-based control.
\newblock In {\em International Conference on Intelligent Robots and Systems},
  pages 5026--5033. IEEE, 2012.

\bibitem[\protect\citeauthoryear{Wan \bgroup \em et al.\egroup
  }{2020}]{wan2020mutual}
Michael Wan, Tanmay Gangwani, and Jian Peng.
\newblock Mutual information based knowledge transfer under state-action
  dimension mismatch.
\newblock In {\em Conference on Uncertainty in Artificial Intelligence}, pages
  1218--1227. PMLR, 2020.

\bibitem[\protect\citeauthoryear{Whitney \bgroup \em et al.\egroup
  }{2020}]{whitney2019dynamicsaware}
William Whitney, Rajat Agarwal, Kyunghyun Cho, and Abhinav Gupta.
\newblock Dynamics-aware embeddings.
\newblock In {\em International Conference on Learning Representations}, 2020.

\bibitem[\protect\citeauthoryear{Wulfmeier \bgroup \em et al.\egroup
  }{2017}]{wulfmeier2017mutual}
Markus Wulfmeier, Ingmar Posner, and Pieter Abbeel.
\newblock Mutual alignment transfer learning.
\newblock {\em arXiv preprint arXiv:1707.07907}, 2017.

\bibitem[\protect\citeauthoryear{Yang \bgroup \em et al.\egroup
  }{2020}]{yang2020efficient}
Tianpei Yang, Jianye Hao, Zhaopeng Meng, Zongzhang Zhang, Yujing Hu, Yingfeng
  Chen, Changjie Fan, Weixun Wang, Zhaodong Wang, and Jiajie Peng.
\newblock Efficient deep reinforcement learning through policy transfer.
\newblock In {\em Proceedings of the 19th International Conference on
  Autonomous Agents and MultiAgent Systems}, pages 2053--2055, 2020.

\bibitem[\protect\citeauthoryear{Zhang \bgroup \em et al.\egroup
  }{2021}]{zhang2021learning}
Qiang Zhang, Tete Xiao, Alexei~A Efros, Lerrel Pinto, and Xiaolong Wang.
\newblock Learning cross-domain correspondence for control with dynamics
  cycle-consistency.
\newblock {\em ICLR}, 2021.

\end{thebibliography}

\begin{appendix}

     \section{Environment Details and Hyperparameters}

     \subsection{Experiment Settings}
     In our experiments, n-step gridworld and Mujoco envrionments are deterministic. Hence, the latent variable $z_t$ is not necessary in the environments and it is adopted only in combat tasks. Without $z_t$, the loss function of the transition model reduces to MSE loss:

     \begin{equation*}
          \label{eq: loss}
          \begin{aligned}
               %_{\tau \sim \pi}
               \mathcal{L}(\theta^D, W^{ae}) = & \mathop{\mathbb{E}}_{s_t, a_t, s_{t + 1}} \big{[} \  ||\tilde{s}_{t+1} - s_{t+1}||_2^2 \big{]}
          \end{aligned}
     \end{equation*}

     \subsection{Gridworld}
\begin{figure*}
          \centering
          \subfigure[Gridworld]{
               \label{fig: task-image:a}
               \includegraphics[height=95pt]{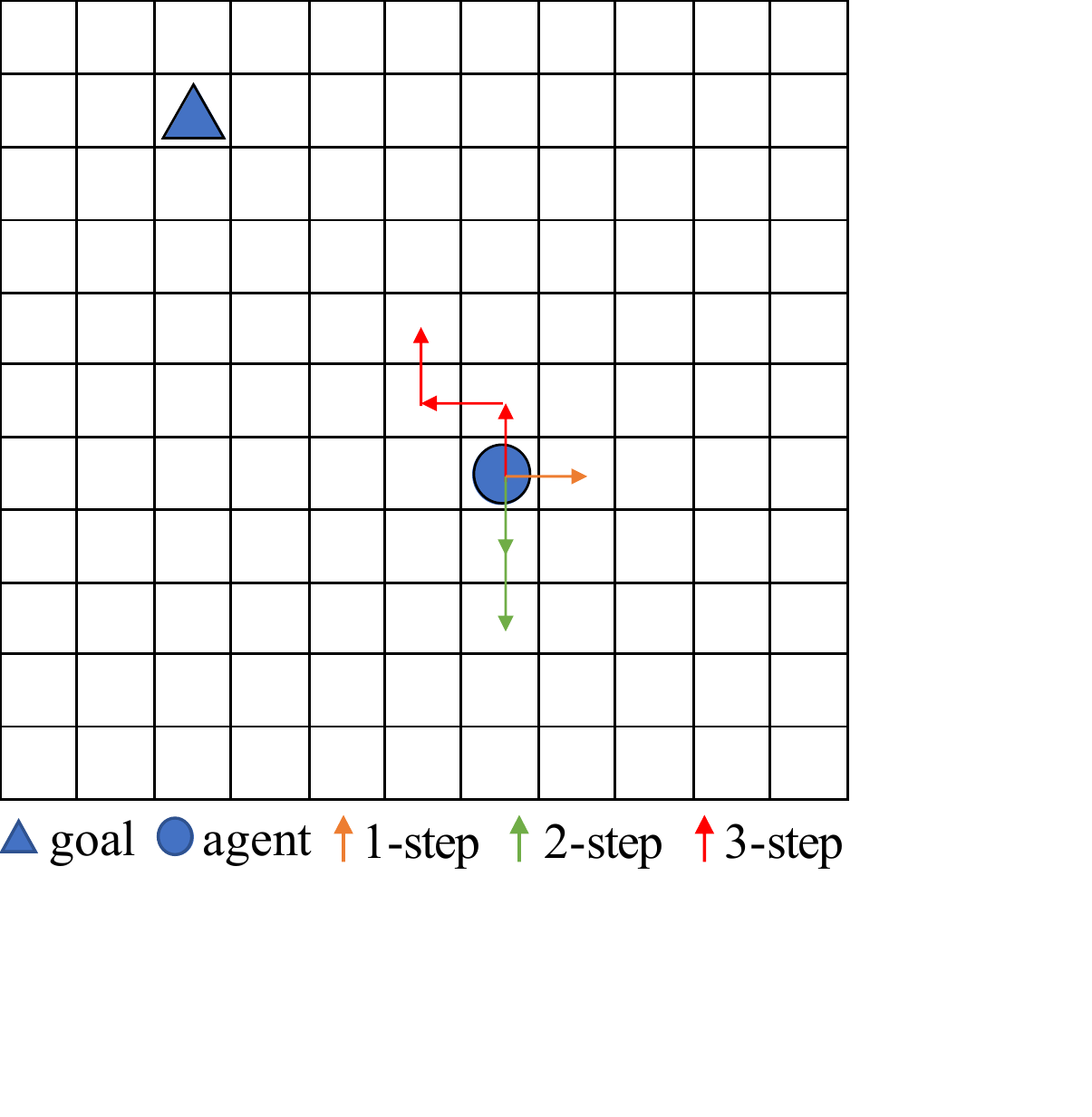}
          } \hspace{-0.5em}
          \subfigure[Pendulum and DoublePendulum]{
               \label{fig: task-image:b}
               \includegraphics[height=95pt]{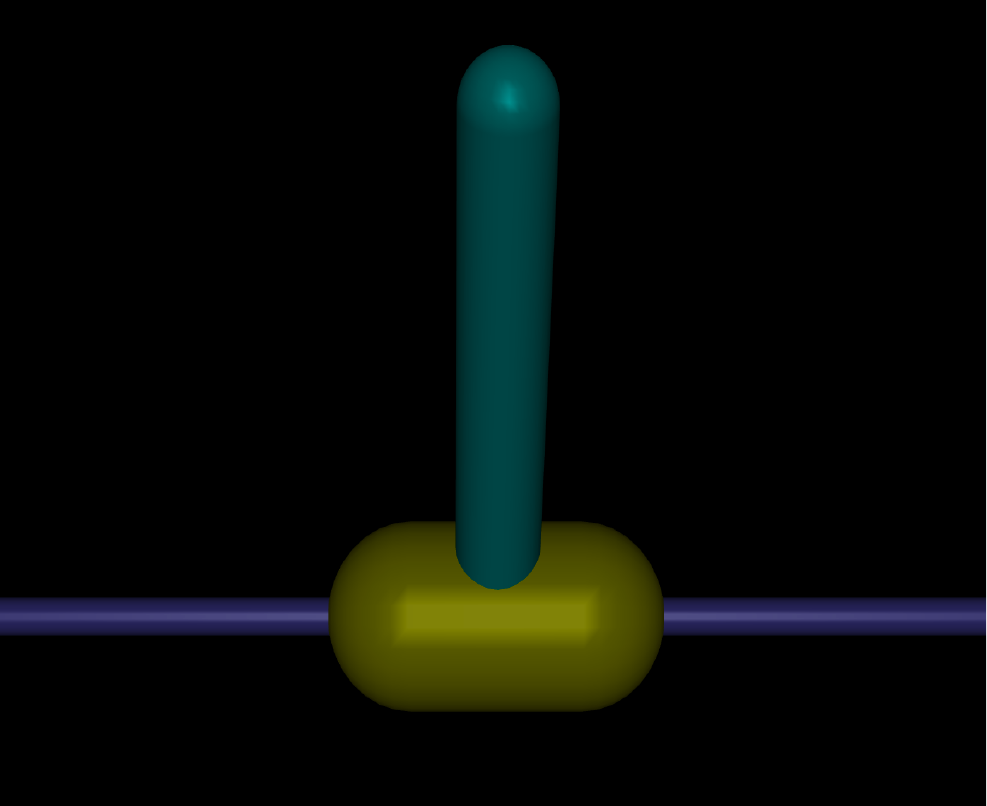}
               \includegraphics[height=95pt]{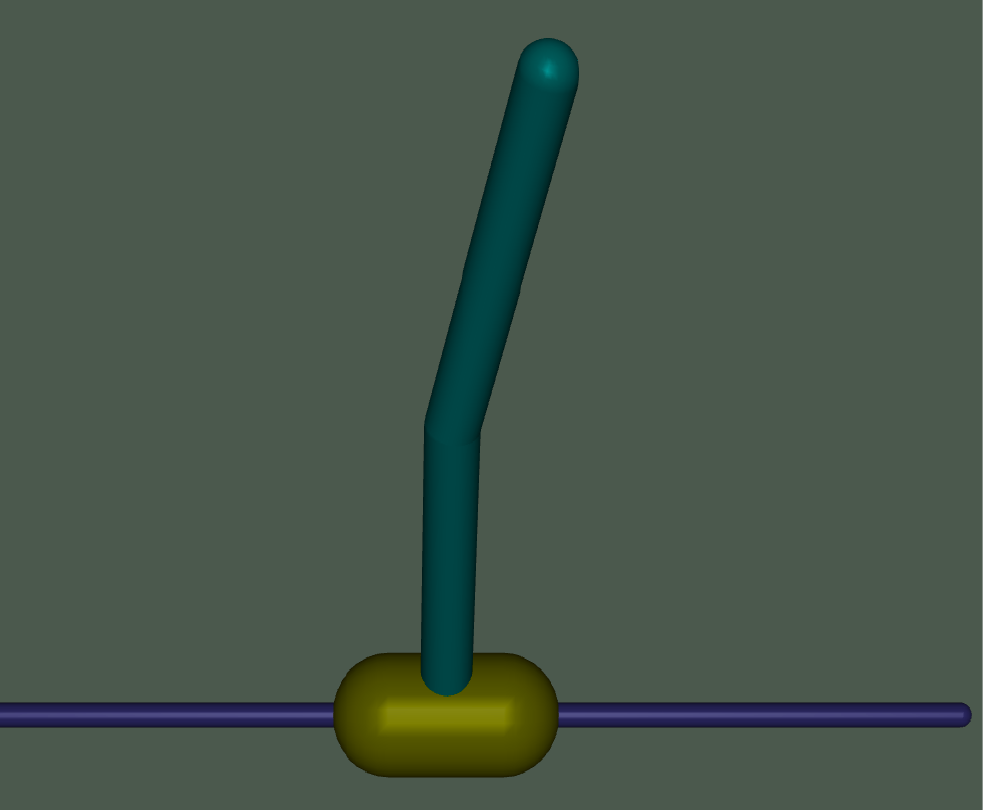}
          } \hspace{-0.5em}
          \subfigure[Combat tasks in the commercial game]{
               \label{fig: task-image:c}
               \includegraphics[height=95pt]{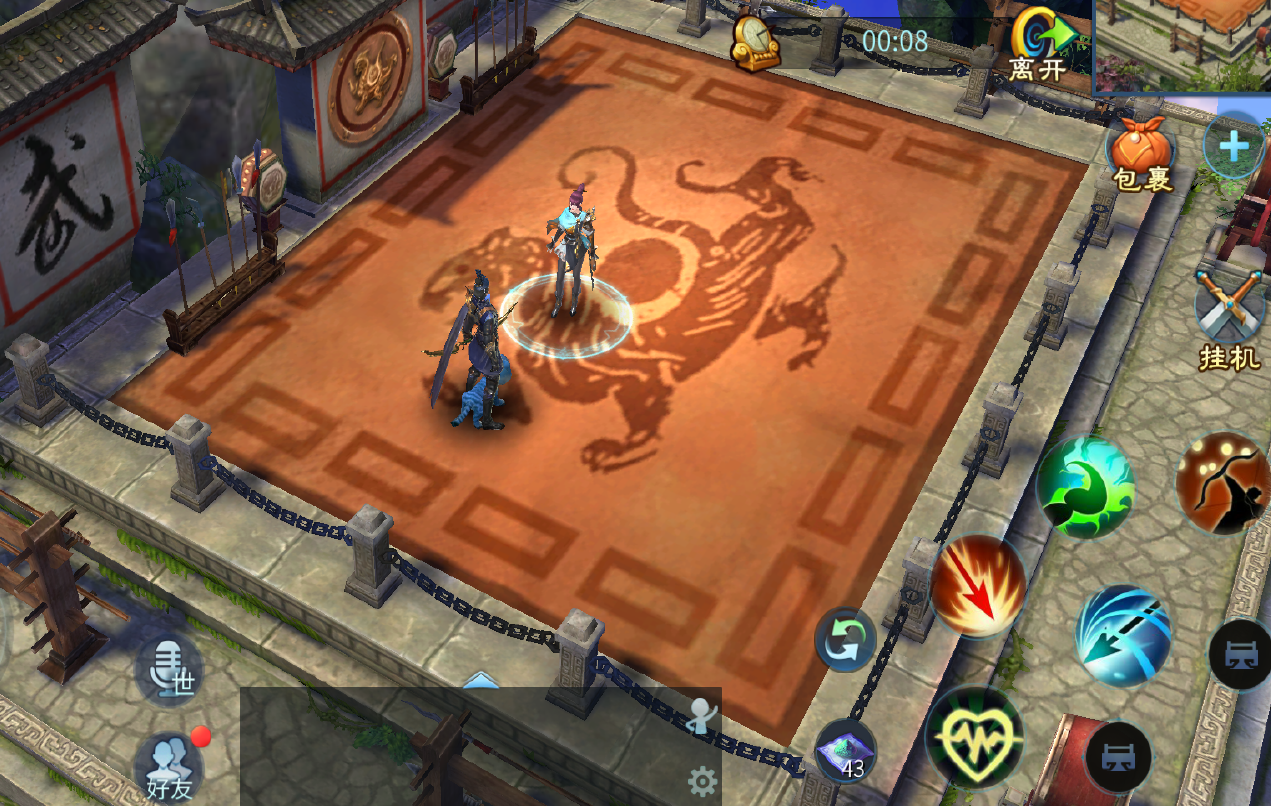}
          }
          \caption{(a) The $n$-step gridworld environment. (b) Pendulum and DoublePendulum tasks. (c) The commercial game.  }
          \label{fig: task-image}
     \end{figure*}

     % In our experiment, we consider $n$-step planning in an $11 \times 11$ gridworld, which is shown in Figure. \ref{fig: grid}.
     % The parameter settings in this experiment is shown in Table. \ref{tab: grid}.  
     Gridworld environment consists of an agent and a goal as shown in Figure \ref{fig: task-image:a}. At each episode, the agent is spawned in a random position and the goal is randomly positioned. The objective of the agent is to reach the goal.

     \textbf{States:} The state is a 4-dimensional vector consisting of the current position $(x, y)$ and the goal position $(\dot{x}, \dot{y})$.

     \textbf{Actions:} An action of the agent in $n$-step gridworld indicates $n$ consecutive moves in four directions. Taking $2$-step gridworld as an example, the action space is 16, including \{ $\uparrow\uparrow, \uparrow\downarrow, \uparrow\leftarrow, \uparrow\rightarrow, \leftarrow\uparrow, \leftarrow\downarrow, \leftarrow\leftarrow, \leftarrow\rightarrow,
          \rightarrow\uparrow, \rightarrow\downarrow, \rightarrow\leftarrow, \rightarrow\rightarrow,
          \downarrow\uparrow, \downarrow\downarrow, \downarrow\leftarrow, \downarrow\rightarrow$ \}.
     Once the agent selects an action, it executes moves step-by-step. If the agent hits the boundary, it will stay in the current position.

     \textbf{Rewards:} The agent receives a -0.05 reward each move and a +10 reward when the agent reaches the goal.

     \textbf{Termination:} Each game is terminated when the agent reaches the goal, or the agent has taken 20 actions.

     The hyperparameters of the experiment are available in Table \ref{tab: grid}.

     \begin{table}[htbp]
          \centering
          \caption{Parameter settings in gridworld experiment}
          \begin{tabular}{c|c|l}
                                                    & Parameters            & \multicolumn{1}{l}{Value}      \\\hline
               \multirow{8}[0]{*}{SAC}              & state\_embed\_dim     & \multicolumn{1}{l}{-}          \\
                                                    & state\_embed\_hiddens & \multicolumn{1}{l}{-}          \\
                                                    & ac hiddens            & \multicolumn{1}{l}{[200, 100]} \\
                                                    & actor lr              & 1e-5                           \\
                                                    & critic lr             & 1e-3                           \\
                                                    & $\tau$                & 0.999                          \\
                                                    & $\alpha$              & 0.2                            \\
                                                    & $\gamma$              & 0.99                           \\\hline
               \multirow{3}[0]{*}{Transition Model} & action\_embed\_dim    & 2                              \\
                                                    & hiddens               & \multicolumn{1}{l}{[64, 32]}   \\
                                                    & lr                    & 1e-3                           \\\hline
          \end{tabular}%
          \label{tab: grid}%
     \end{table}%

     \subsection{Mujoco and Roboschool}
     % The tasks in Mujoco and Roboschool is illustrated in Figure. \ref{fig: mujoco}.
     we conduct experiments on \emph{InvertedPendulum} and \emph{InvertedDoublePendulum} tasks in two simulator, \textbf{Mujoco} and \textbf{Roboschool}, denoted by \emph{mP}, \emph{rP}, \emph{mDP} and \emph{rDP}, respectively. In all tasks, the agent needs to apply a force to the cart to balance the poles upright. Figure \ref{fig: task-image:b} presents the screenshots of the tasks.

     \textbf{States:} The state of \emph{mP} is a 4-dimensional vector, representing the position and velocity of the cart and the pole. For \emph{rP}, the state is a 5-dimensional vector, including the position and velocity of the cart and the joint angle and its sine and cosine.
     The state vector of \emph{mDP} consists of the cart's position , sine and cosine of two joint angles, velocities of position and the angles, and constraint forces on position and the angles, with a total length of 11. For \emph{rDP}, the state is described via the position and velocity of the cart, pole's position, two angles, and sine and cosine of the angles.

     \textbf{Actions:} We discretize original continuous action spaces into equally spaced values. We summarize the detailed configurations in Table \ref{tab: task}.

     \textbf{Rewards:} In \emph{mP} and \emph{rP}, the agent gets +1 reward for keeping the pole upright. For \emph{mDP} and \emph{rDP}, the agent receives +10 reward each step and a negative reward for hight velocity and drift from the center point.

     \textbf{Termination:} Each game is terminated when the agent has taken 100 actions, or the poles are too low.

     Table \ref{tab: mujoco} shows the hyperparameters in this experiment.

     \begin{table}[htbp]
          \centering
          \caption{Configurations of the four environments}
          \begin{tabular}{cccc}
               \hline
               Task & State Dim & Act Range & Discretized Act Dim \\
               \hline
               mP   & 4         & [-3, 3]   & 101                 \\
               mDP  & 11        & [-1, 1]   & 51                  \\
               rP   & 5         & [-1, 1]   & 91                  \\
               rDP  & 9         & [-1, 1]   & 71                  \\
               \hline
          \end{tabular}%
          \label{tab: task}%
     \end{table}%

     \begin{table}[htbp]
          \centering
          \caption{Parameter settings in Mujoco and Roboschool experiment}
          \begin{tabular}{c|c|l}
                                                    & Parameters            & \multicolumn{1}{l}{Value}      \\\hline
               \multirow{8}[0]{*}{SAC}              & state\_embed\_dim     & \multicolumn{1}{l}{5}          \\
                                                    & state\_embed\_hiddens & \multicolumn{1}{l}{[200, 100]} \\
                                                    & ac\_hiddens           & \multicolumn{1}{l}{[200, 100]} \\
                                                    & actor\_lr             & 1e-5                           \\
                                                    & critic\_lr            & 1e-3                           \\
                                                    & $\tau$                & 0.999                          \\
                                                    & $\alpha$              & 0.2                            \\
                                                    & $\gamma$              & 0.99                           \\\hline
               \multirow{3}[0]{*}{Transition Model} & action\_embed\_dim    & 3                              \\
                                                    & hiddens               & \multicolumn{1}{l}{[64, 32]}   \\
                                                    & lr                    & 1e-3                           \\\hline
          \end{tabular}%
          \label{tab: mujoco}%
     \end{table}%

     \subsection{Combat Tasks in the Commercial Game}

     \begin{figure*}
          \centering
          \includegraphics[height=15pt]{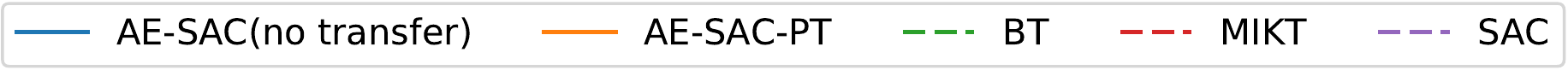}
          \vspace{-0.7em}

          \subfigure[task $n=1$ (source task $n=2$)]{
               \includegraphics[width=0.25\textwidth]{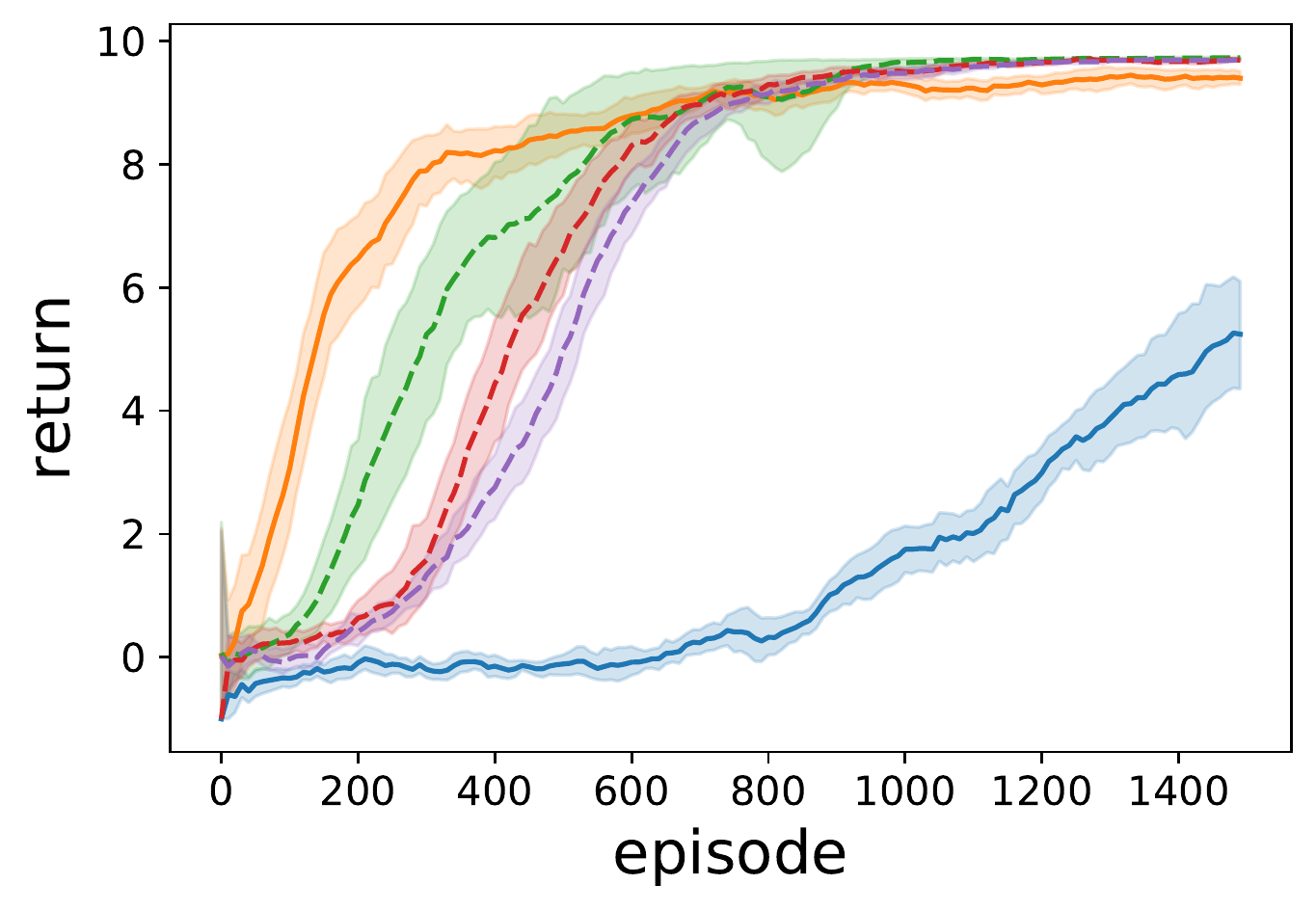}
          }
          % \subfigure[$n=1$(source task $n=3$)]{
          %     \includegraphics[width=0.22\textwidth]{fig/gridworld-2-0.pdf}
          % } \hspace{-1em}
          % \subfigure[$n=2$(source task $n=1$)]{
          %     \includegraphics[width=0.22\textwidth]{fig/gridworld-0-1.pdf}
          % }
          \subfigure[task $n=2$ (source task $n=3$)]{
               \includegraphics[width=0.25\textwidth]{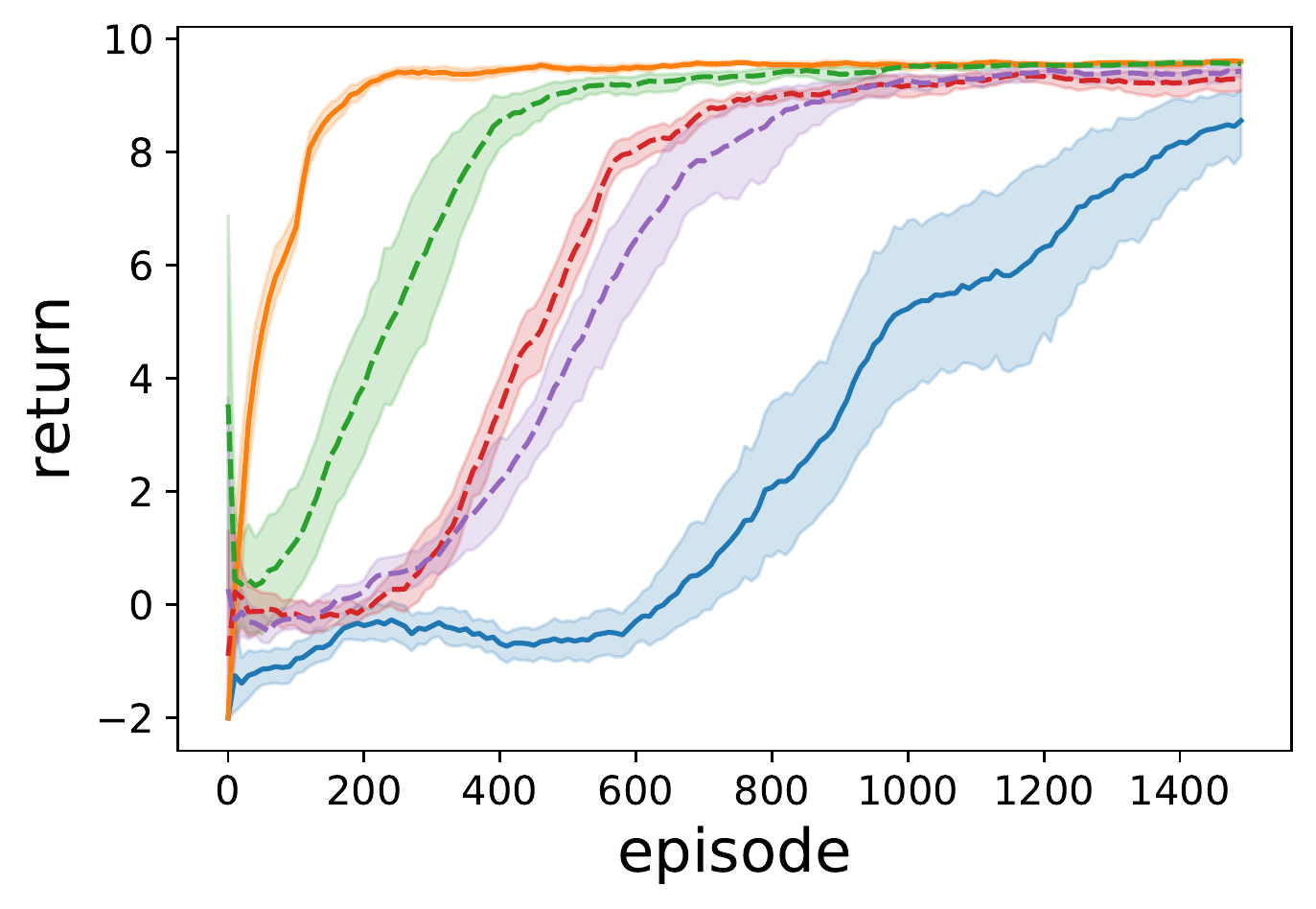}
          }
          \subfigure[task $n=3$ (source task $n=1$)]{
               \includegraphics[width=0.25\textwidth]{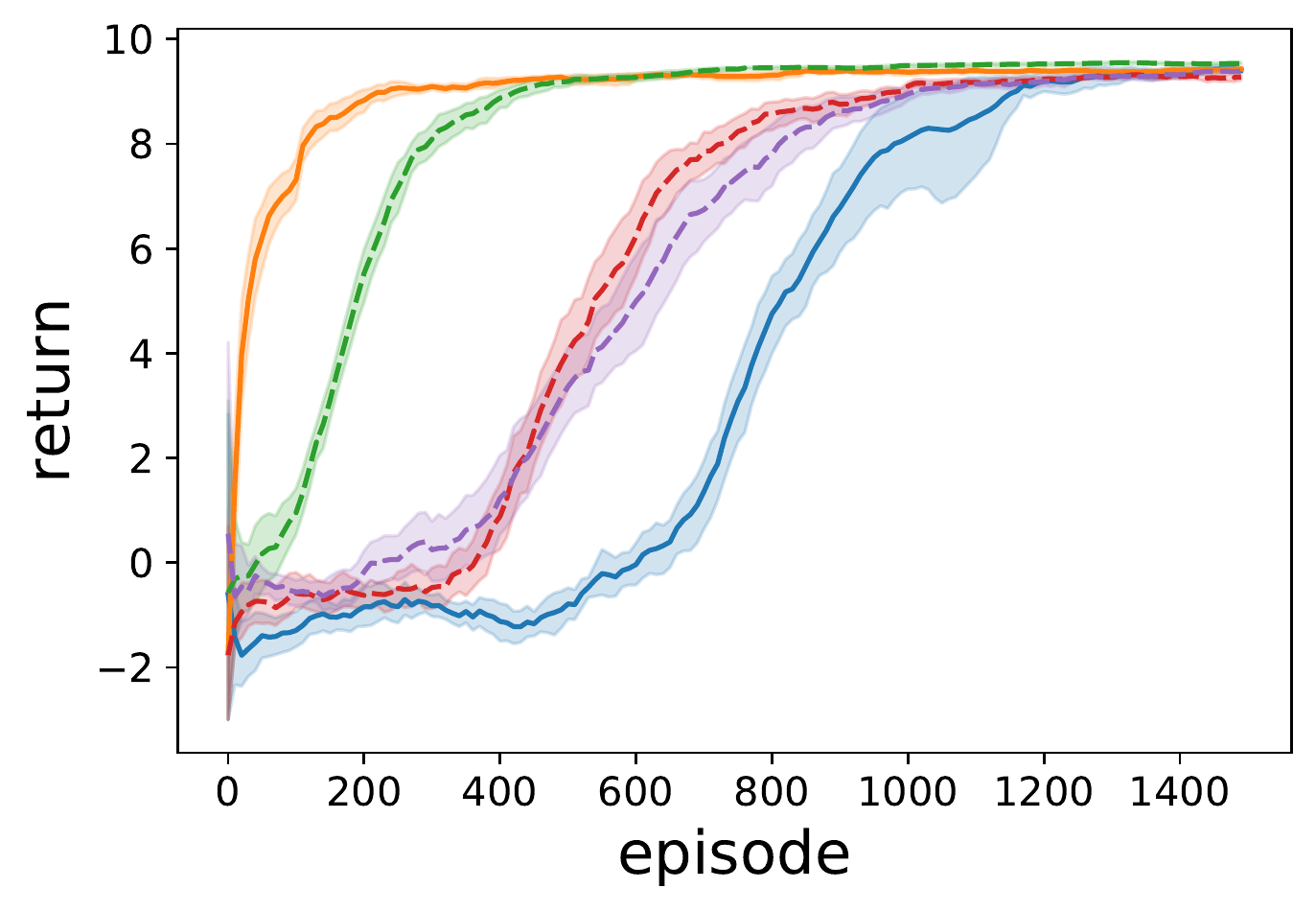}
          }
          % \subfigure[$n=3$(source task $n=2$)]{
          %     \includegraphics[width=0.22\textwidth]{fig/gridworld-1-2.pdf}
          % }
          \caption{The experiment results on Gridworld. The solid lines denote our method, and the dashed lines represent the compared methods. The shaded areas are bootstrapped 95\% confidence intervals.}
          \label{fig: gridworld-result}
     \end{figure*}

     In the scenario, the agent needs to move and arrange the skills reasonably to defeat an opponent controlled by rules. Figure \ref{fig: task-image:c} displays the screenshot of the scenario.
     To have a better understanding of learned action embeddings in the paper, we summarize the effects of \emph{She Shou}'s skills in Table \ref{tab: skill}.

     \begin{table}[htbp]
          \centering
          \caption{Skill Descriptions of \emph{She Shou}}
          \resizebox{0.5\textwidth}{70pt}{
               \begin{tabular}{lll}
                    Skill ID & effect                                                  \\\hline
                    1        & Cause damage and fire dot damage for 15 seconds         \\\hline
                    2        & Cause damage and poisoning damage for 7 seconds         \\\hline
                    3        & Cause damage and poisoning damage for 7 seconds         \\\hline
                    4        & Set a trap and cause damage every second                \\\hline
                    5        & Cause damage and fire dot damage for 15 seconds         \\\hline
                    6        & Cause damage and poisoning damage for 7 seconds         \\\hline
                    7        & Cause damage                                            \\\hline
                    8        & Stun enemy for 2.5 seconds                              \\\hline
                    9        & Summon an arrow tower, which causes damage every second \\\hline
                    10       & Cause damage                                            \\\hline
                    11       & Cause damage                                            \\\hline
                    12       & Knock back enemy                                        \\\hline
                    13       & Cause damage and silence enemy for 6 seconds            \\\hline
                    14       & Cause large amount of damage and create an aura         \\\hline
               \end{tabular}%
          }
          \label{tab: skill}%
     \end{table}%

     \textbf{States:} The state consists of the HPs, attack, defense, and buffs of the agent and the opponent, and whether skills are available. Thus, different classes have different sizes of the state vector.

     \textbf{Actions:} The action space contains six unique skills of the class and some common operations, such as move, attack, and dummy.

     \textbf{Rewards:} Each step, the reward is calculated by the HP change of the agent and the opponent. Besides, a negative reward is given if the agent selects an unready skill. A final reward +10 is given if the agent wins, -10 otherwise.

     \textbf{Termination:} Each game is terminated when the agent has taken 100 actions, or anyone of both sides is dead.

     The hyperparameters used in this experiment are available in Table \ref{tab: qnyh}.

     \begin{table}[htbp]
          \centering
          \caption{Parameter settings in combat tasks experiment}
          \begin{tabular}{c|c|l}
                                                    & Parameters            & \multicolumn{1}{l}{Value}      \\\hline
               \multirow{8}[0]{*}{SAC}              & state\_embed\_dim     & \multicolumn{1}{l}{25}         \\
                                                    & state\_embed\_hiddens & \multicolumn{1}{l}{[200, 100]} \\
                                                    & ac\_hiddens           & \multicolumn{1}{l}{[300, 200]} \\
                                                    & actor\_lr             & 1e-5                           \\
                                                    & critic\_lr            & 1e-3                           \\
                                                    & $\tau$                & 0.999                          \\
                                                    & $\alpha$              & 0.2                            \\
                                                    & $\gamma$              & 0.99                           \\\hline
               \multirow{6}[0]{*}{Transition Model} & action\_embed\_dim    & 6                              \\
                                                    & hiddens               & \multicolumn{1}{l}{[128, 64]}  \\
                                                    & $z$\_dim              & 8                              \\
                                                    & $z$\_hiddens          & [32,]                          \\
                                                    & $\beta$               & 1e-2                           \\
                                                    & lr                    & 1e-3                           \\\hline
          \end{tabular}%
          \label{tab: qnyh}%
     \end{table}%

     \section{Extended Results}

     \begin{figure*}
          \centering
          \includegraphics[height=15pt]{fig/appendix-legend.pdf}
          \vspace{-0.5em}

          \subfigure[\textit{mP} (source task \textit{rP})]{
               \includegraphics[width=0.24\textwidth]{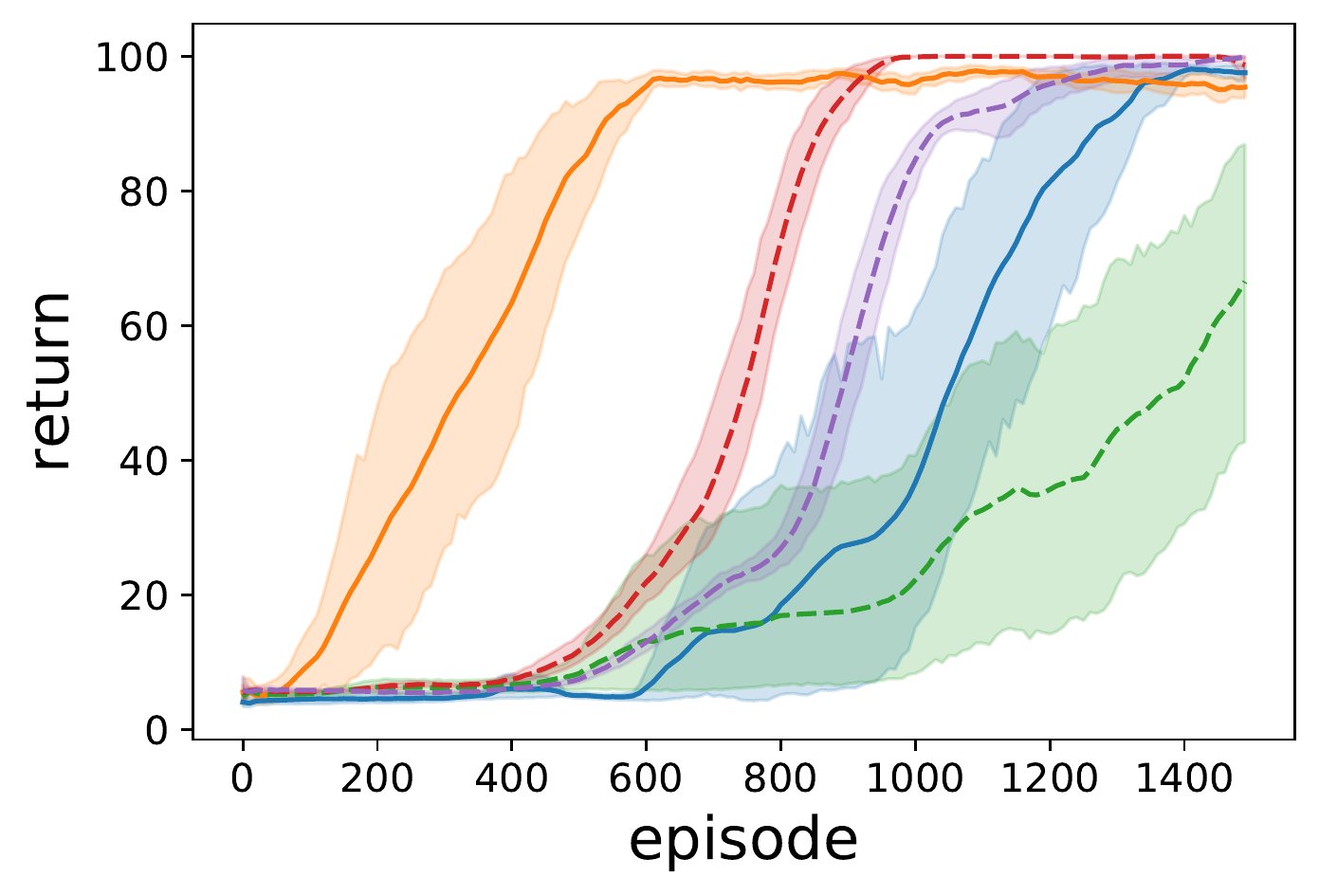}
          } \hspace{-1em}
          % \subfigure[\textit{mP}(source task \textit{mDP})]{
          %     \includegraphics[width=0.22\textwidth]{fig/mujoco-2-0.pdf}
          % } \hspace{-1em}
          \subfigure[\textit{mP} (source task \textit{rDP})]{
               \includegraphics[width=0.24\textwidth]{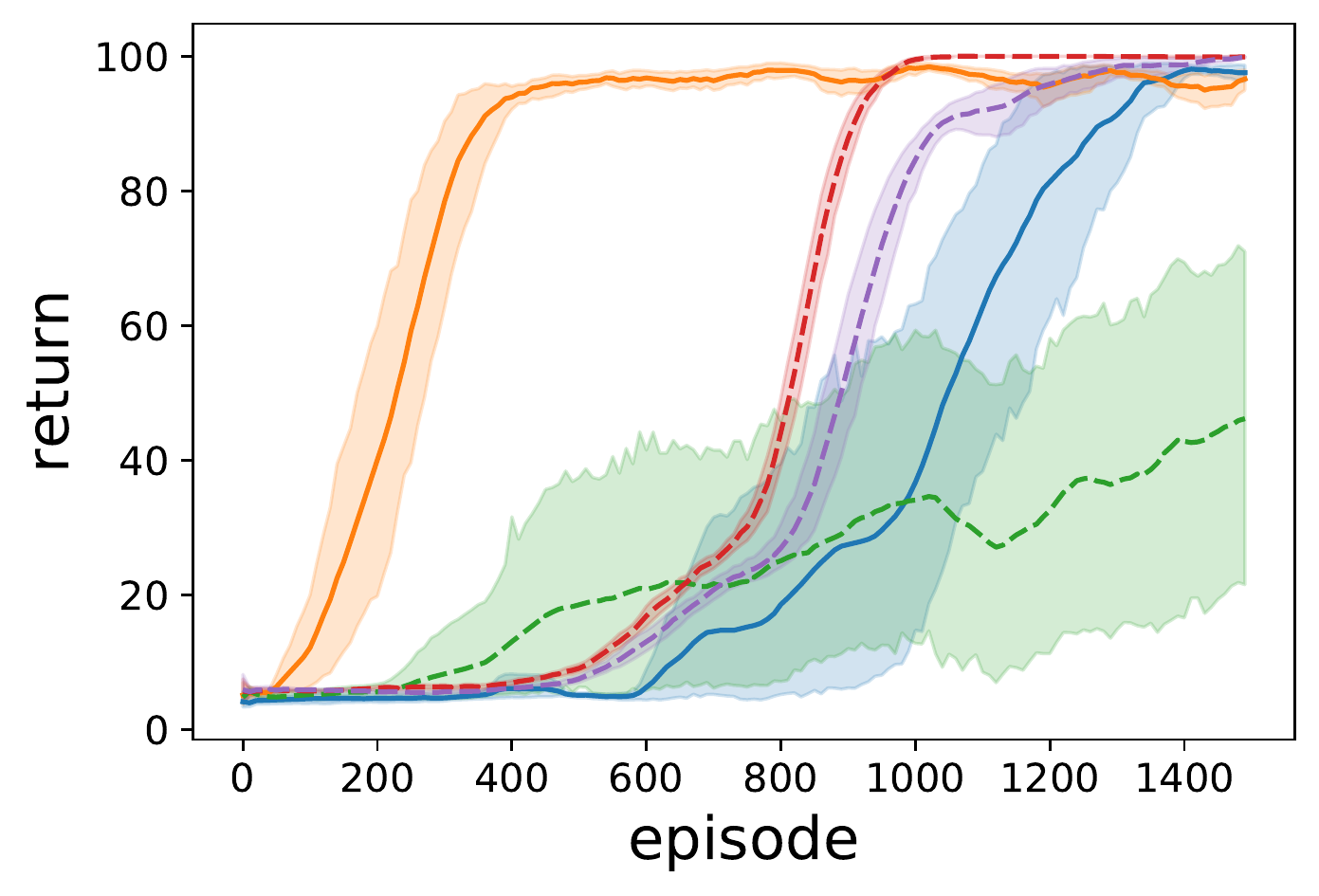}
          } \hspace{-1em}
          \subfigure[\textit{rP} (source task \textit{mP})]{
               \includegraphics[width=0.24\textwidth]{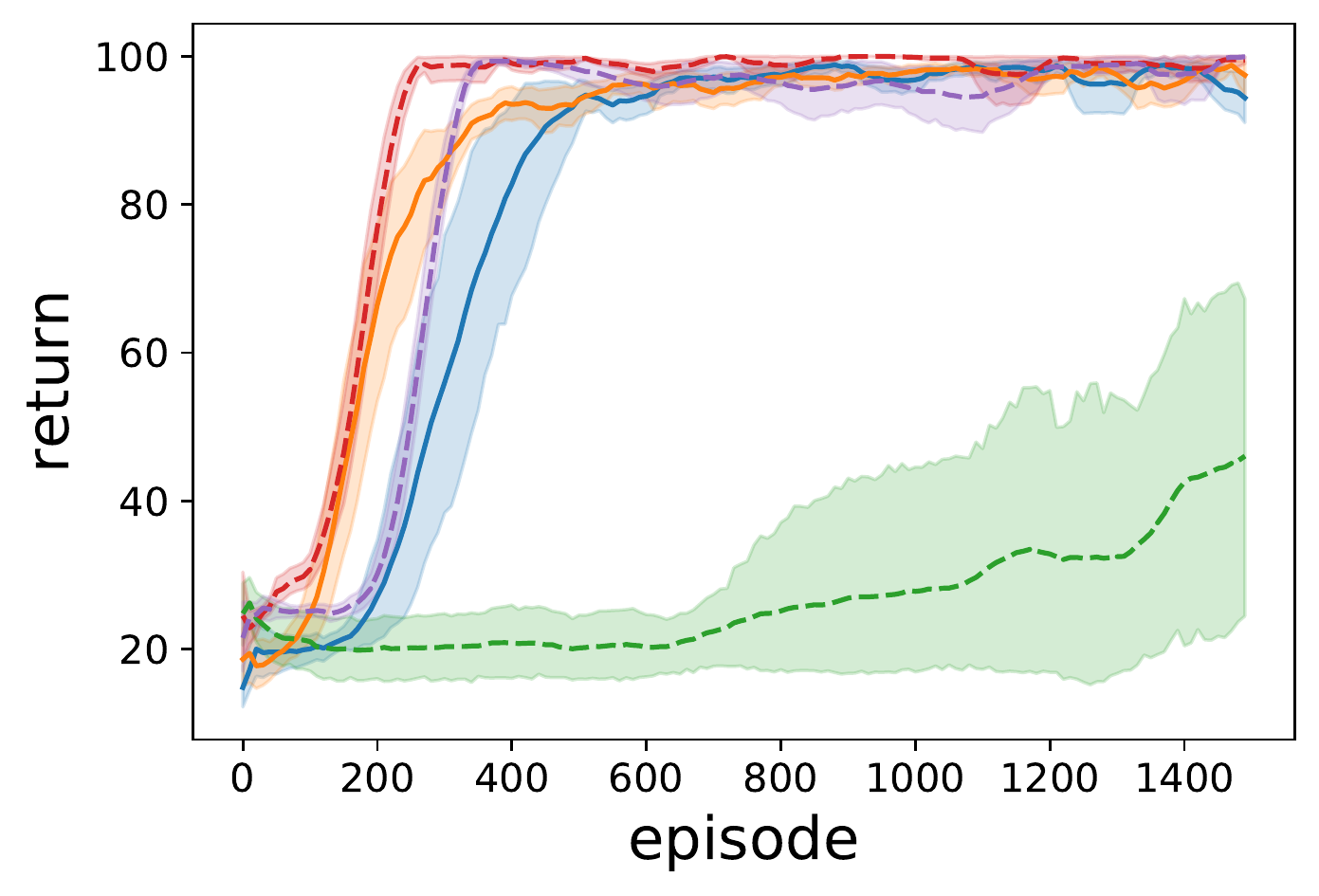}
          } \hspace{-1em}
          \subfigure[\textit{rP} (source task \textit{mDP})]{
               \includegraphics[width=0.24\textwidth]{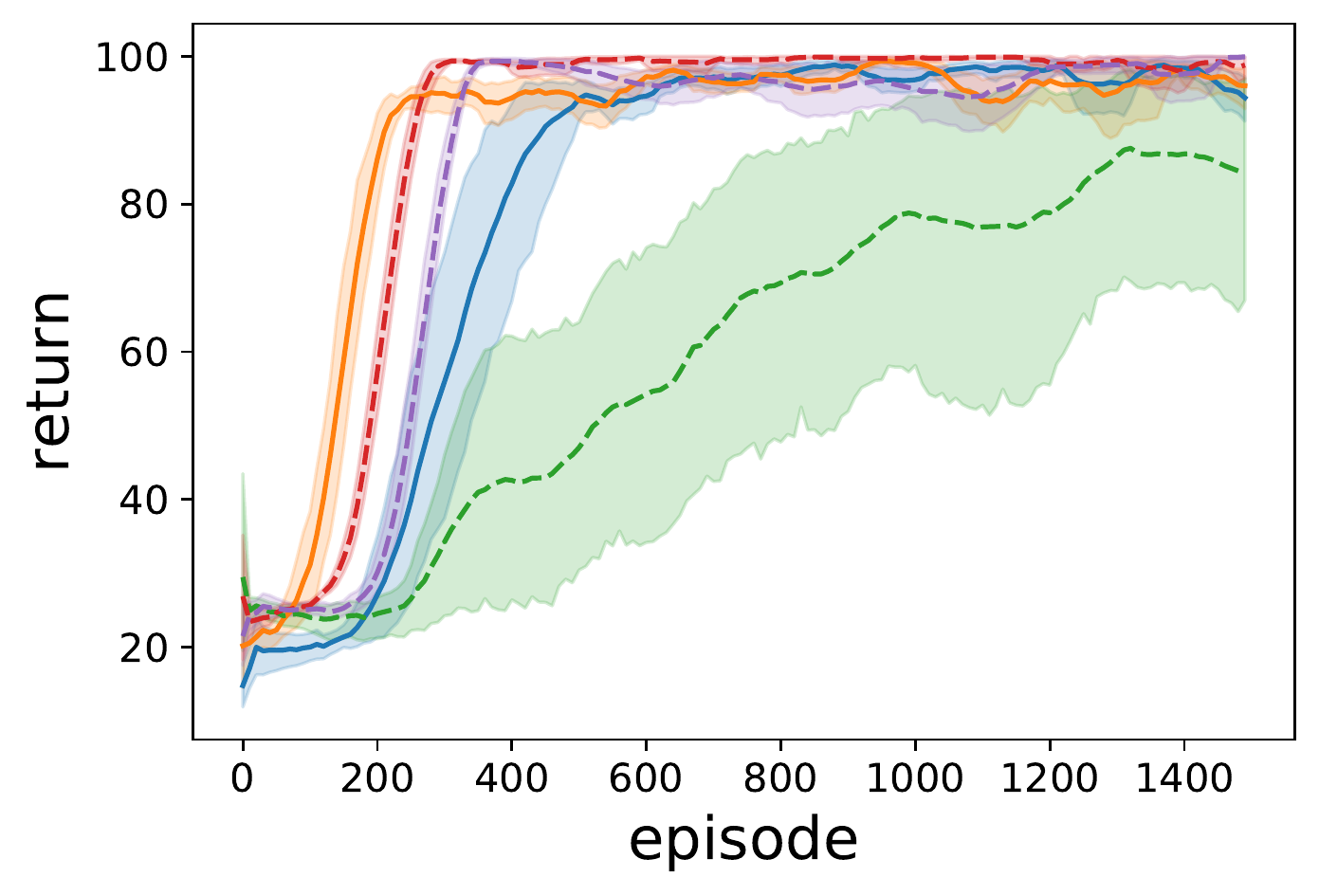}
          }

          % \subfigure[\textit{rP}(source task \textit{rDP})]{
          %     \includegraphics[width=0.22\textwidth]{fig/mujoco-3-1.pdf}
          % }

          \subfigure[\textit{mDP} (source task \textit{mP})]{
               \includegraphics[width=0.24\textwidth]{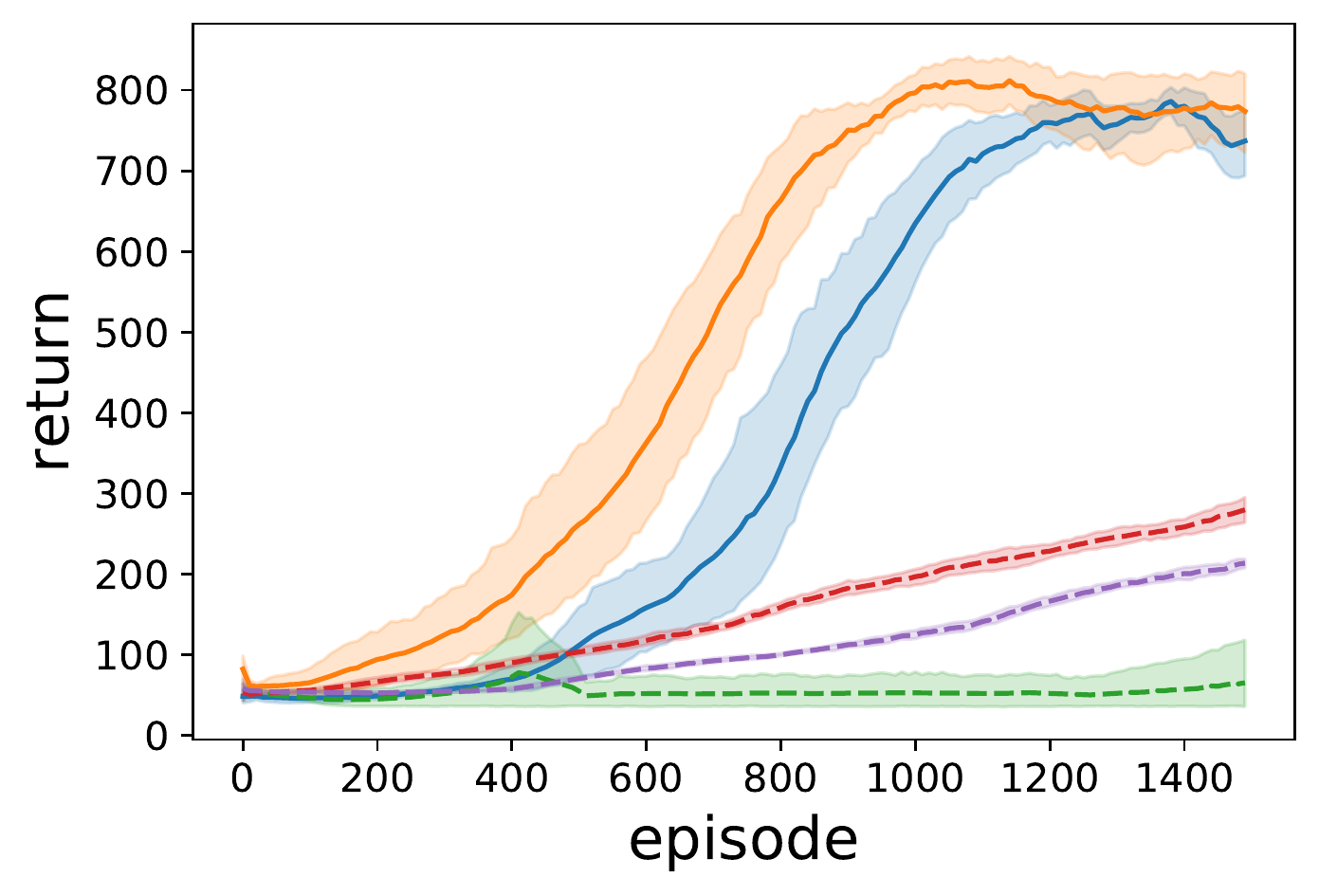}
          } \hspace{-1em}
          % \subfigure[\textit{mDP}(source task \textit{rP})]{
          %     \includegraphics[width=0.22\textwidth]{fig/mujoco-1-2.pdf}
          % } \hspace{-1em}
          \subfigure[\textit{mDP} (source task \textit{rDP})]{
               \includegraphics[width=0.24\textwidth]{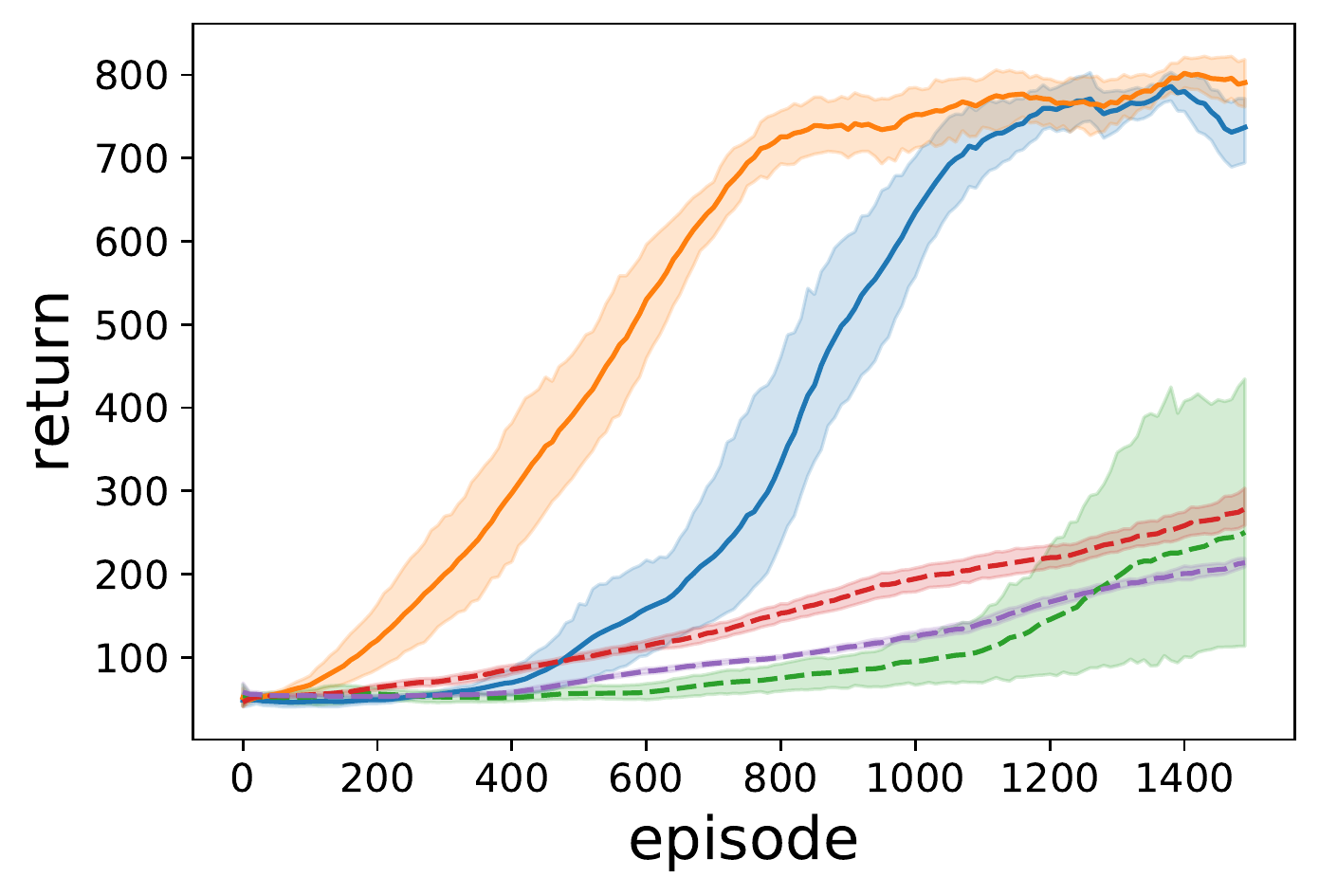}
          } \hspace{-1em}
          % \subfigure[\textit{rDP}(source task \textit{mP})]{
          %     \includegraphics[width=0.22\textwidth]{fig/mujoco-0-3.pdf}
          % } \hspace{-1em}
          \subfigure[\textit{rDP} (source task \textit{rP})]{
               \includegraphics[width=0.24\textwidth]{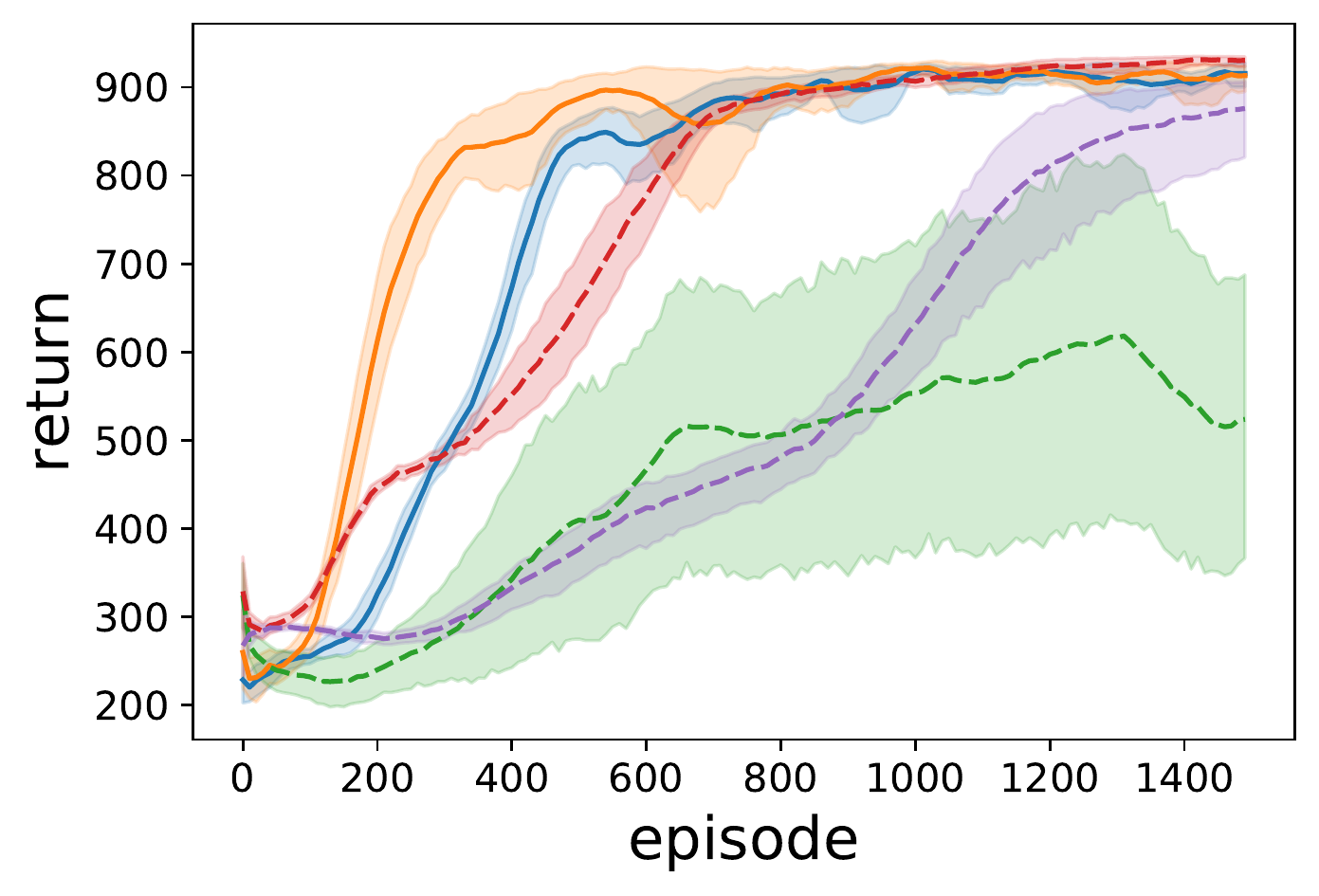}
          } \hspace{-1em}
          \subfigure[\textit{rDP} (source task \textit{mDP})]{
               \includegraphics[width=0.24\textwidth]{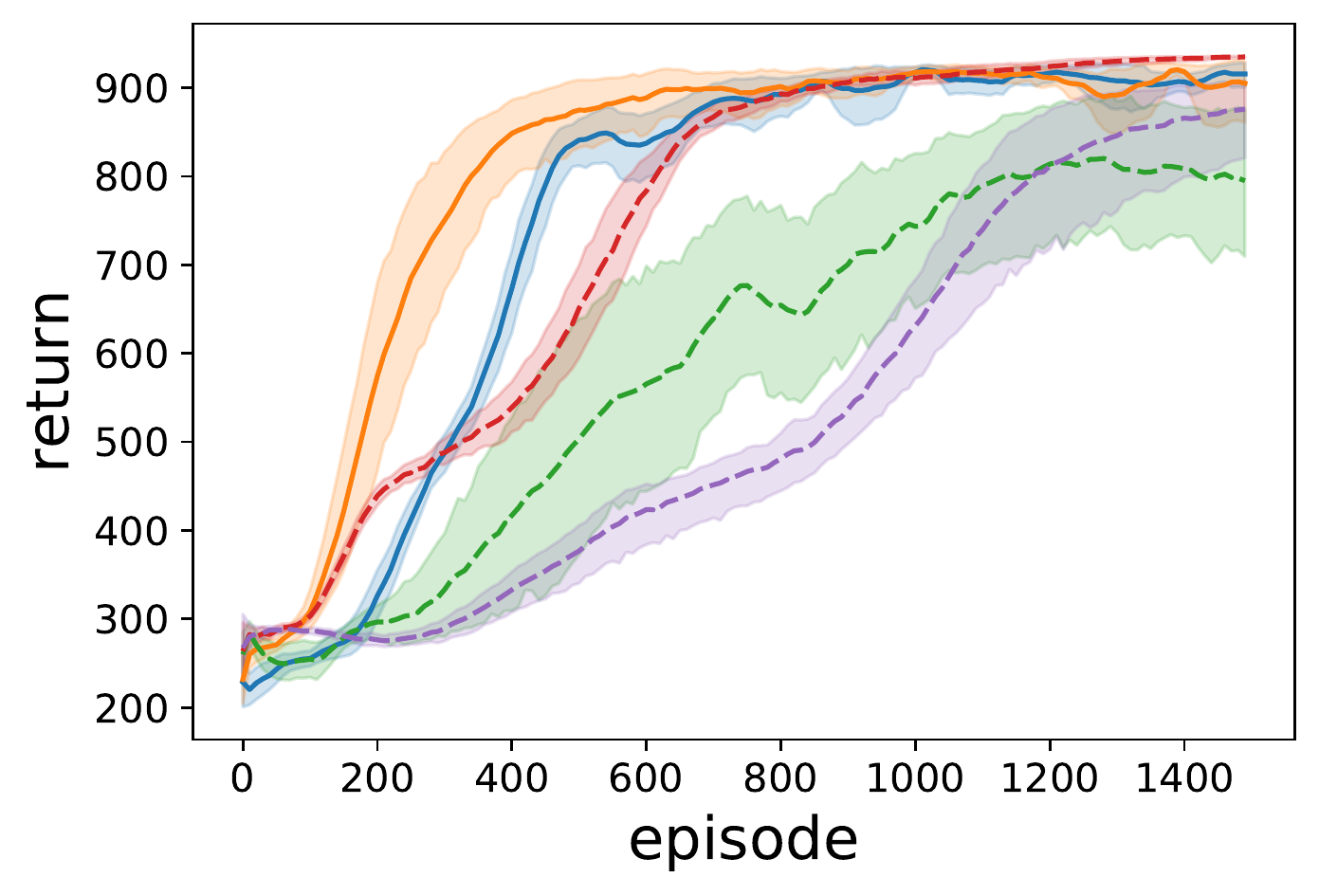}
          }
          \caption{The experiment results on Mujoco and Roboschool. The solid lines denote our method, and the dashed lines represent the compared methods.}
          \label{fig: mujoco-result}
     \end{figure*}

     \begin{figure}
          \centering
          \includegraphics[width=150pt]{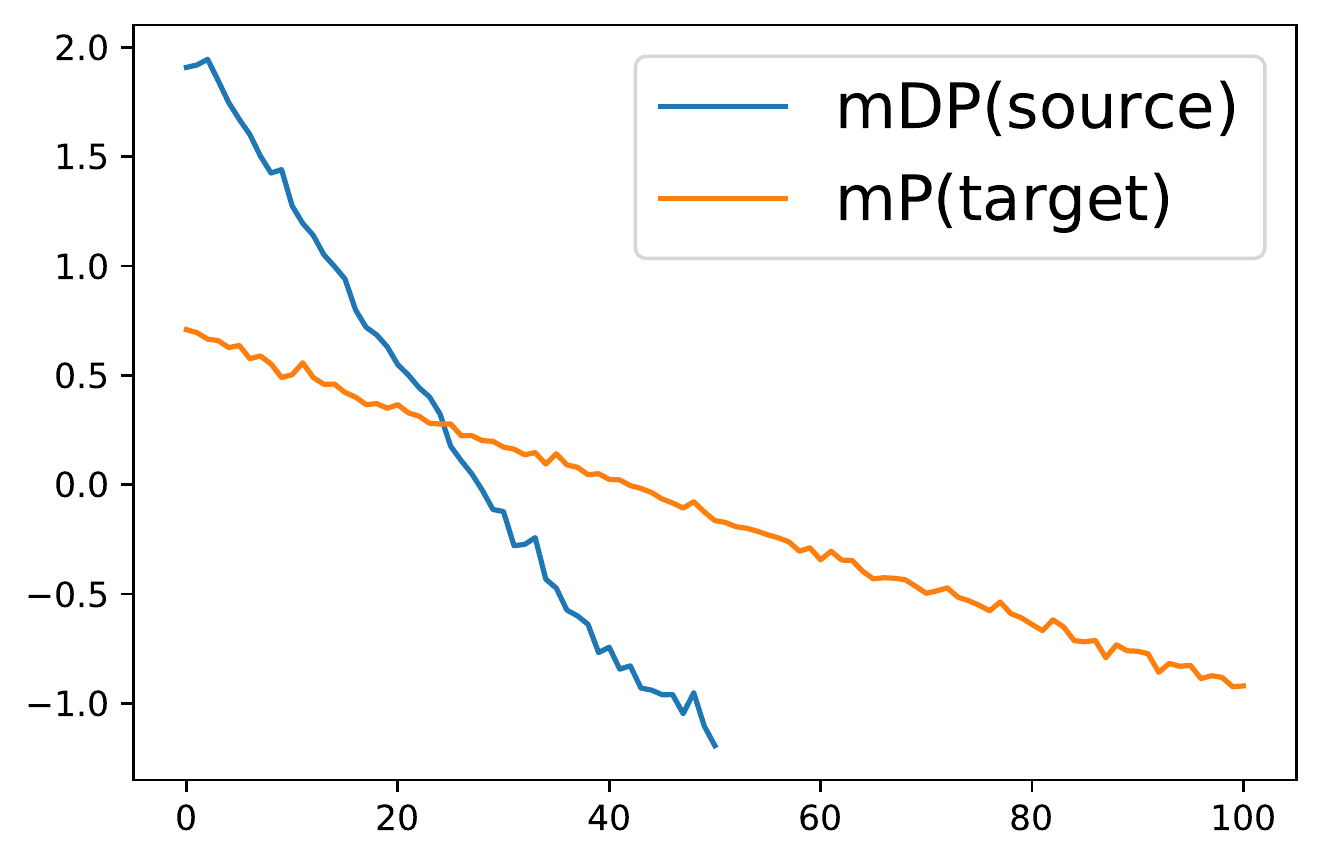}
          \caption{Action embeddings (projected via PCA) of the source task (\emph{mDP}, colored in orange) and the target task (\emph{mP}, colored in blue). The x-axis denotes the index of discretized actions, and the y-axis is its corresponding projected value.}
          \label{fig: mujoco-embedding}
     \end{figure}

     \subsection{Action Embedding}
     In our paper, we have shown the learned embeddings of gridworld and combat tasks. Here, we exhibit action embeddings learned from additional state embeddings in Mujoco and Roboschool tasks.
     The relation between actions should be linear in the tasks since they are discretized from a continuous action space. Figure \ref{fig: mujoco-embedding} plots the PCA projections (1 dimensional) of embeddings. As seen, though there are some local oscillations, the overall trend of the curve is linear. It proves that our method can still learn meaningful latent action representations based on state embeddings.

     \subsection{Policy Transfer}

     For each set of environments, there are many different combinations of source and target tasks. Here we report the rest transfer results of our method from different source tasks that are not included in the paper.

     Figure \ref{fig: gridworld-result} and Figure \ref{fig: mujoco-result} show the results on different tasks. As shown in figures, AE-SAC-PT outperforms BT and MIKT in all transfer tasks and exhibits improved sample efficiency compared to SAC. Moreover, we can see that BT leads to negative transfer in most cases. In contrast, AE-SAC-PT can handle all the transfer tasks and accelerate learning.

\begin{figure}
          \centering
          \includegraphics[width=240pt]{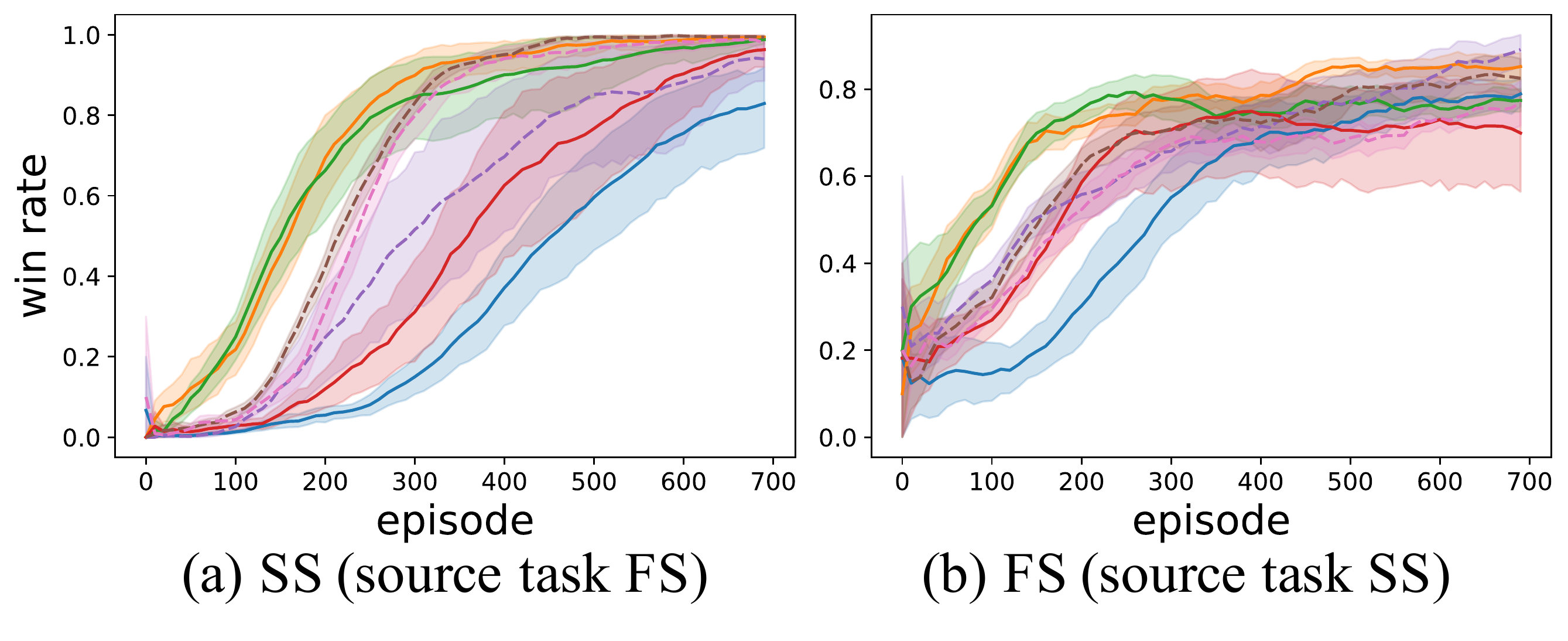}
          %  \caption{The learning curves of methods on the fighting video game. }
          %  \label{fig: qnyh-result}
          %   \subfigure[]{
          %       \label{fig: qnyh-result:a}
          %       \includegraphics[width=115pt]{fig/qnyh-sheshou-new.pdf}
          %   } \hspace{-1em}
          %   \subfigure[]{
          %     \label{fig: qnyh-result:b}
          %     \includegraphics[width=115pt]{fig/qnyh-fangshi-new.pdf}
          %   }
          \caption{Experiment results on combat tasks. The legend refers to the paper.}
          \label{fig: qnyh-res}
     \end{figure}

     Figure \ref{fig: qnyh-res} depicts the full experiment results on the combat tasks.

     \begin{figure*}
          \centering
          \subfigure[$d=2$]{
               \includegraphics[width=0.18\textwidth]{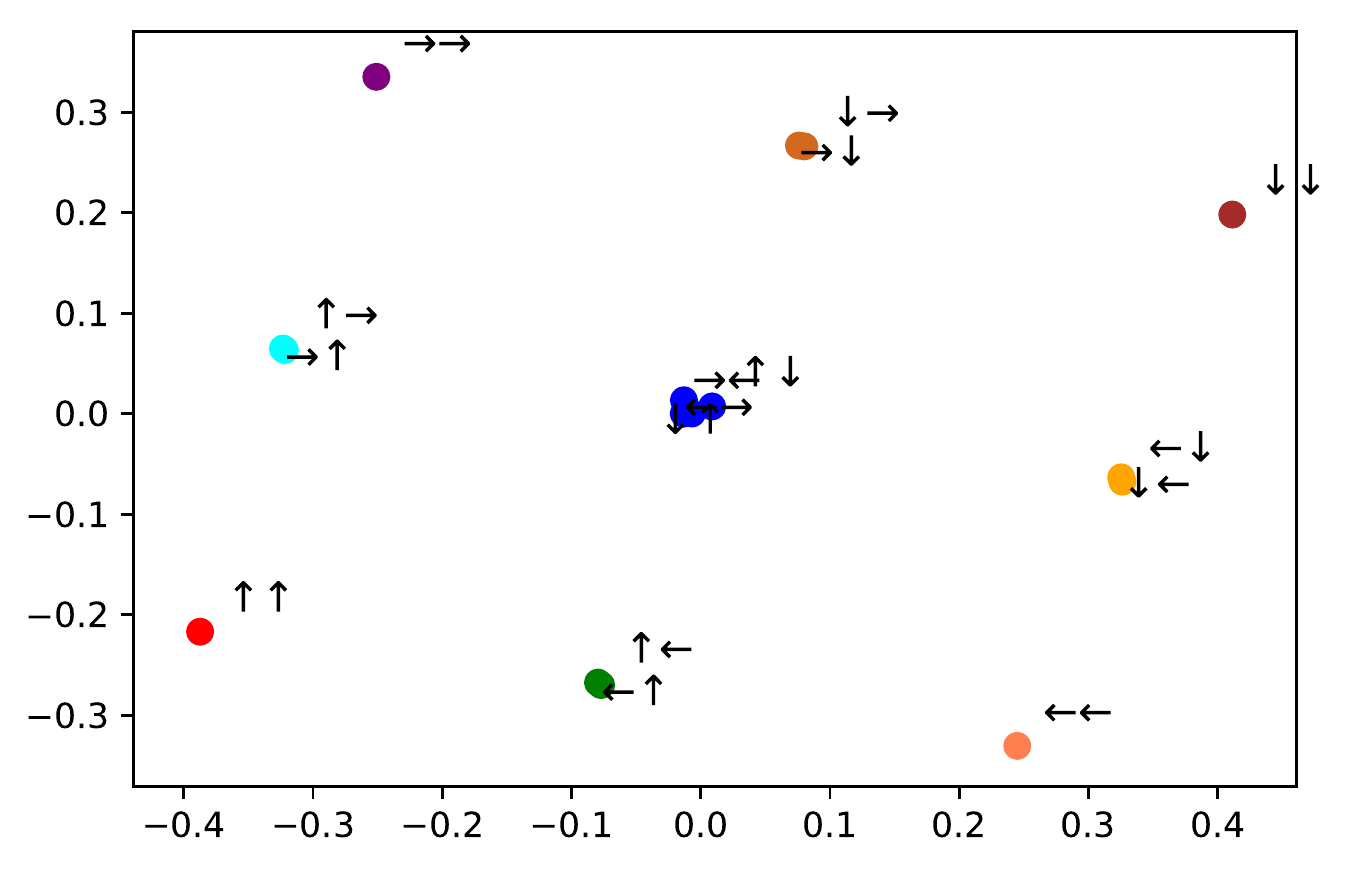}
          }
          \subfigure[$d=3$]{
               \includegraphics[width=0.18\textwidth]{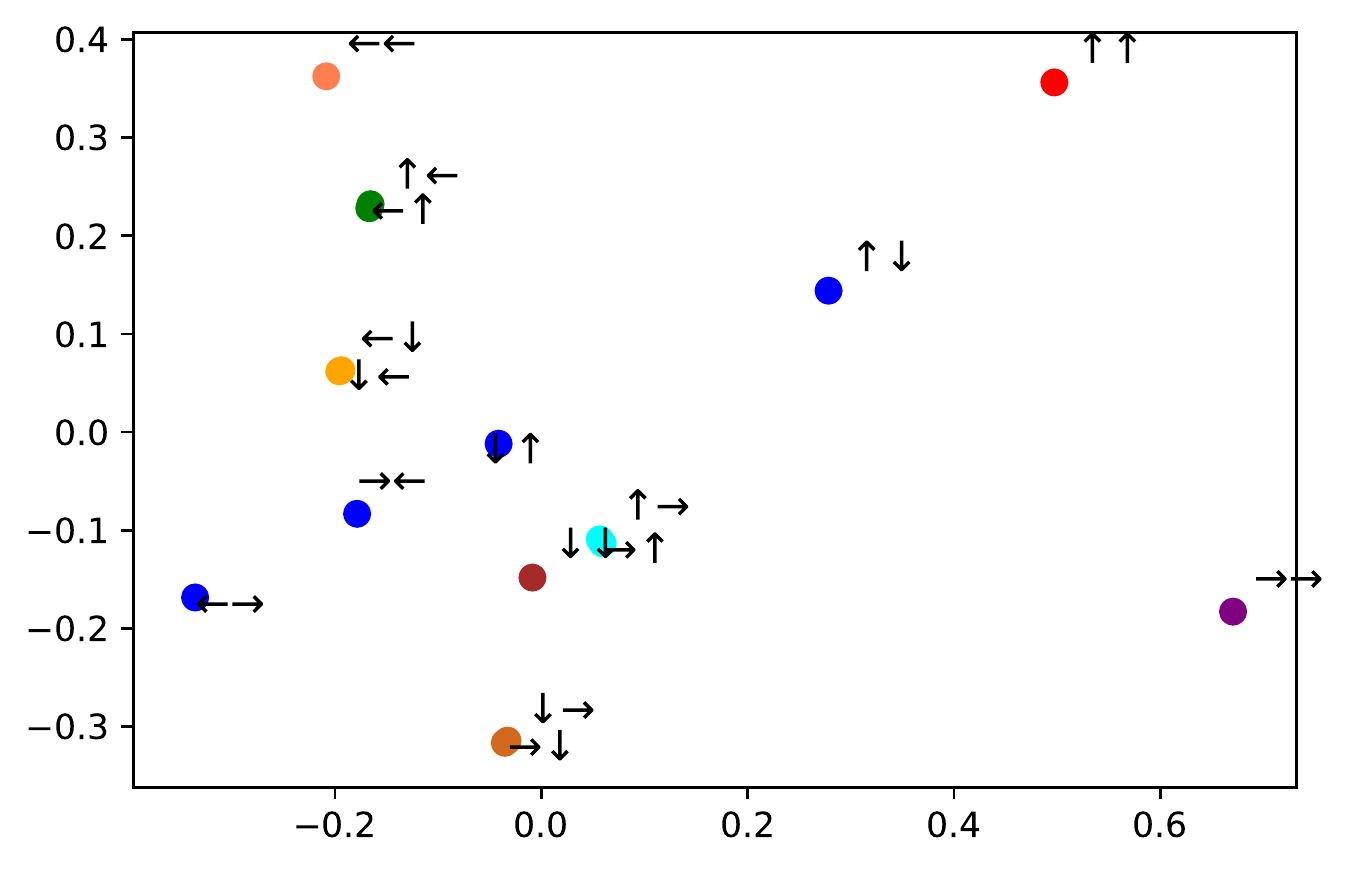}
          }
          \subfigure[$d=4$]{
               \includegraphics[width=0.18\textwidth]{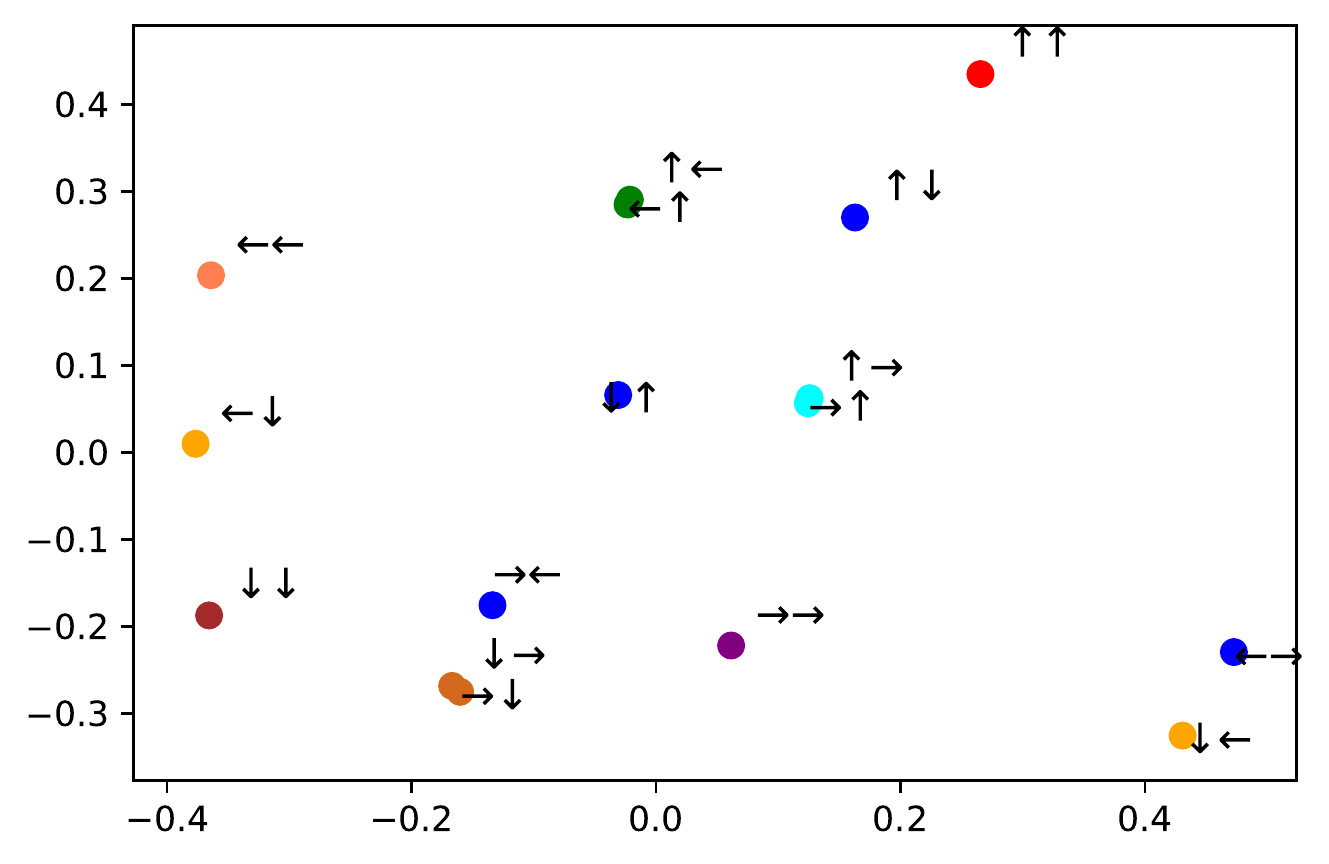}
          }
          \subfigure[$d=5$]{
               \includegraphics[width=0.18\textwidth]{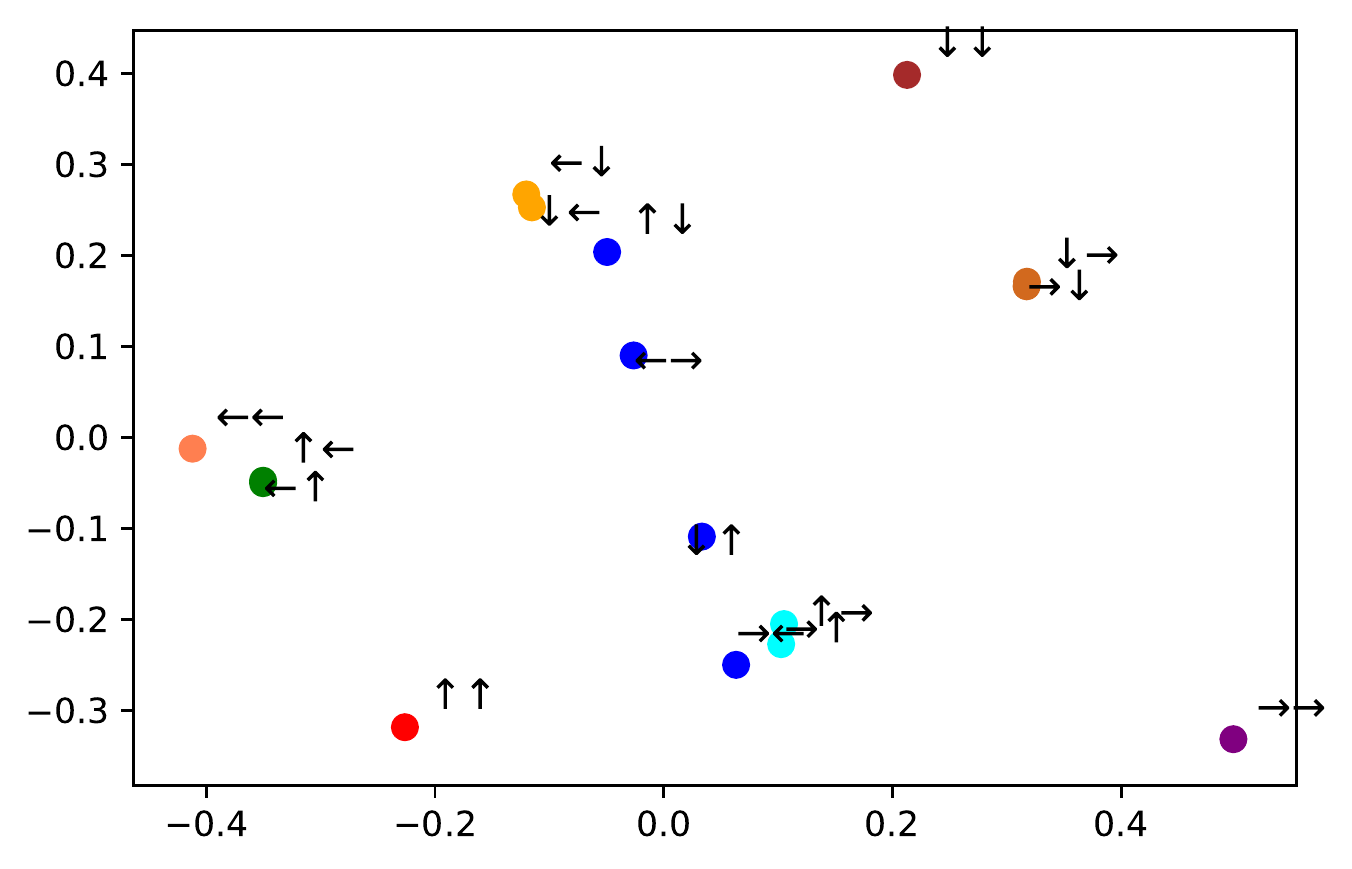}
          }
          \subfigure[$d=6$]{
               \includegraphics[width=0.18\textwidth]{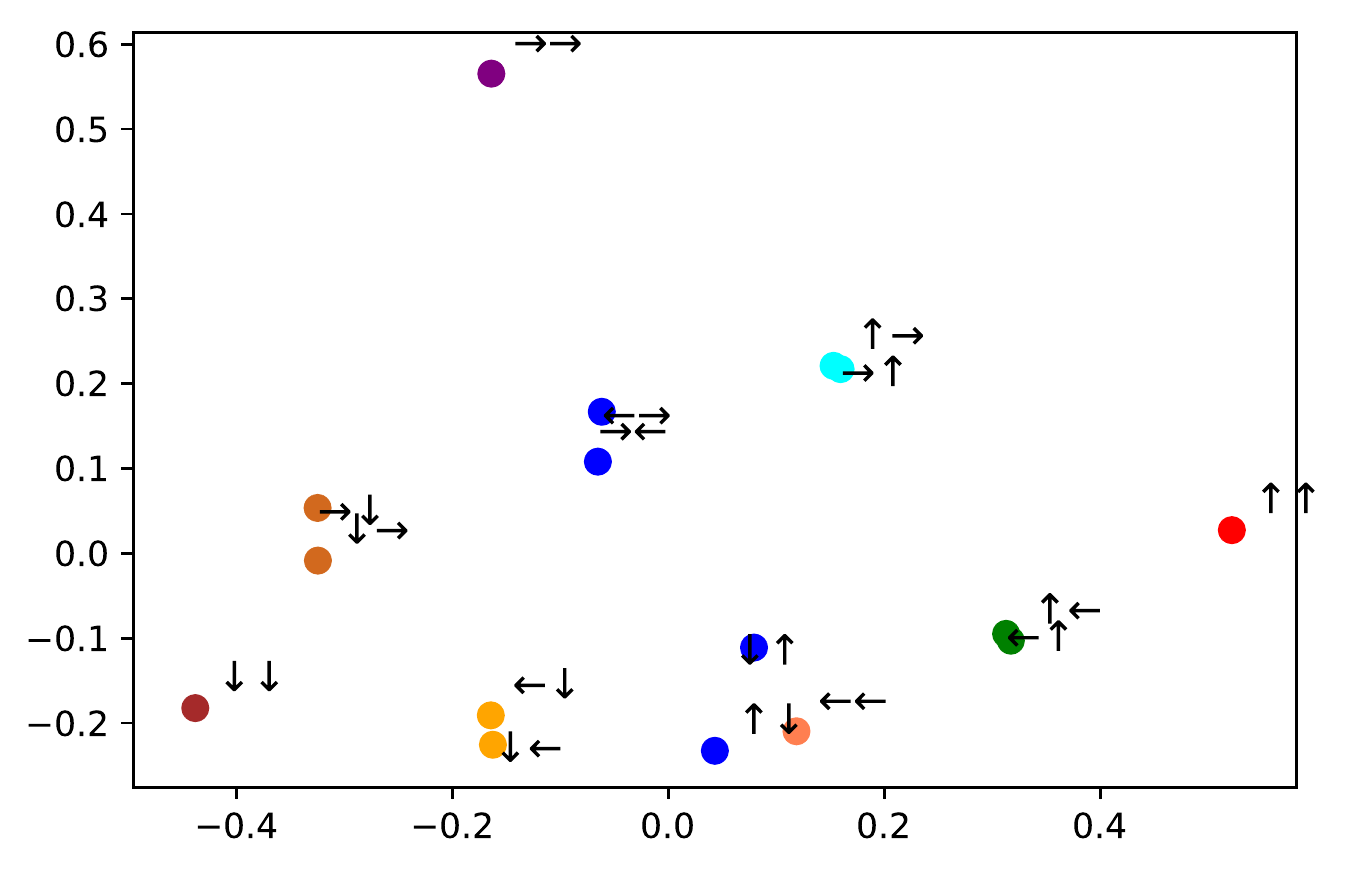}
          }
          \caption{Visualizations of learned action embeddings with different dimensions.}
          \label{fig: d-embedding}
     \end{figure*}

     \begin{figure*}
          \centering
          \includegraphics[height=15pt]{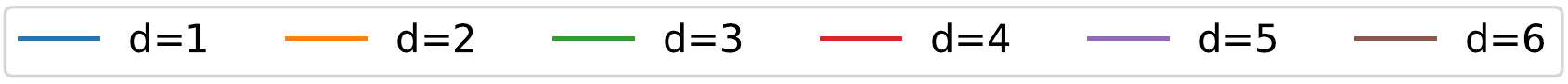}
          \vspace{-0.7em}

          % \subfigure[$n=1$(without transfer)]{
          %     \includegraphics[width=0.3\textwidth]{fig/appendix-dim-seq1.pdf}
          % } \hspace{-1em}
          \subfigure[without transfer]{
               \includegraphics[width=0.25\textwidth]{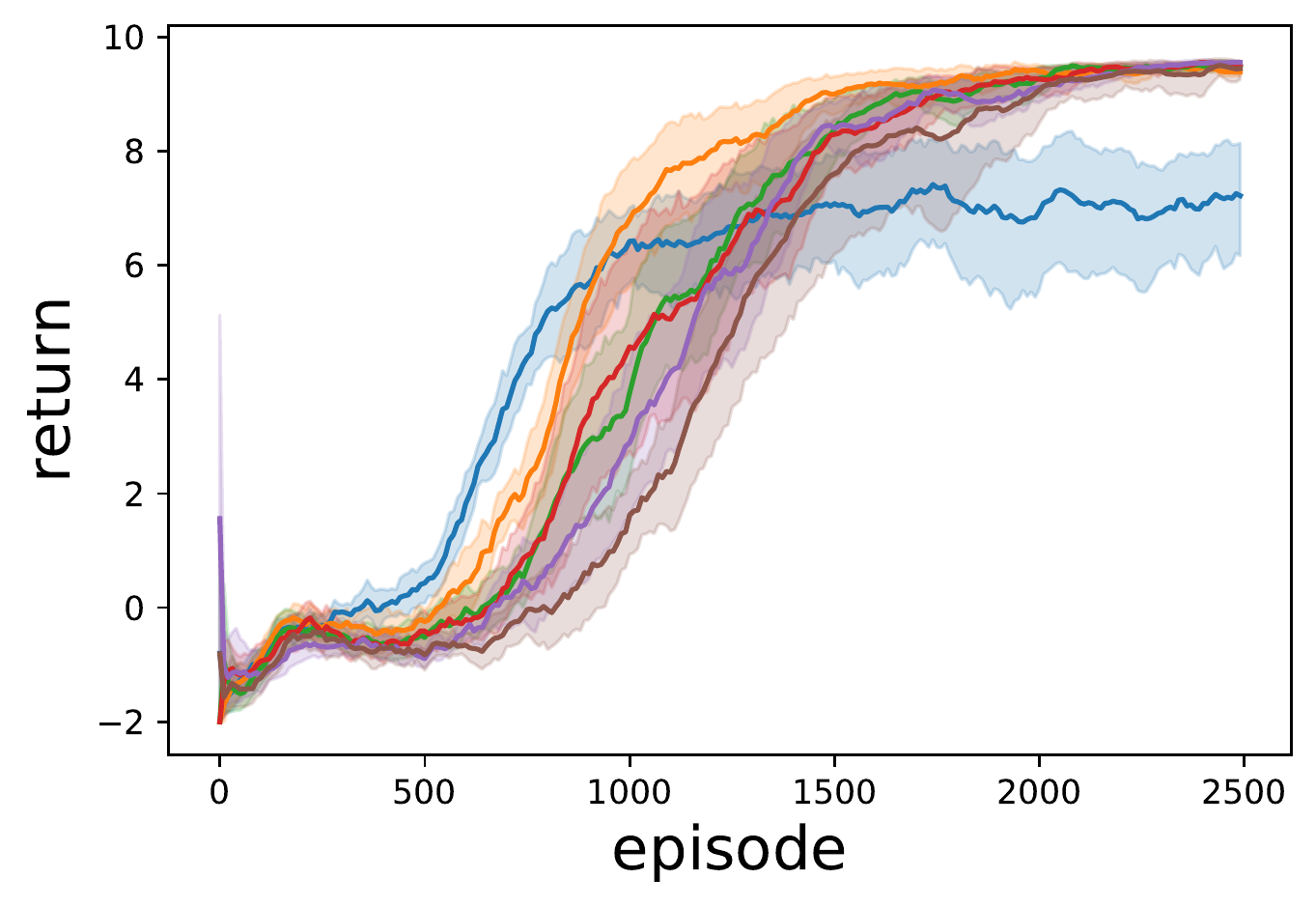}
          }
          % \subfigure[$n=3$(without transfer)]{
          %     \includegraphics[width=0.3\textwidth]{fig/appendix-dim-seq3.pdf}
          % } \vspace{-1em}
          \subfigure[transfer from 1-step gridworld]{
               \includegraphics[width=0.25\textwidth]{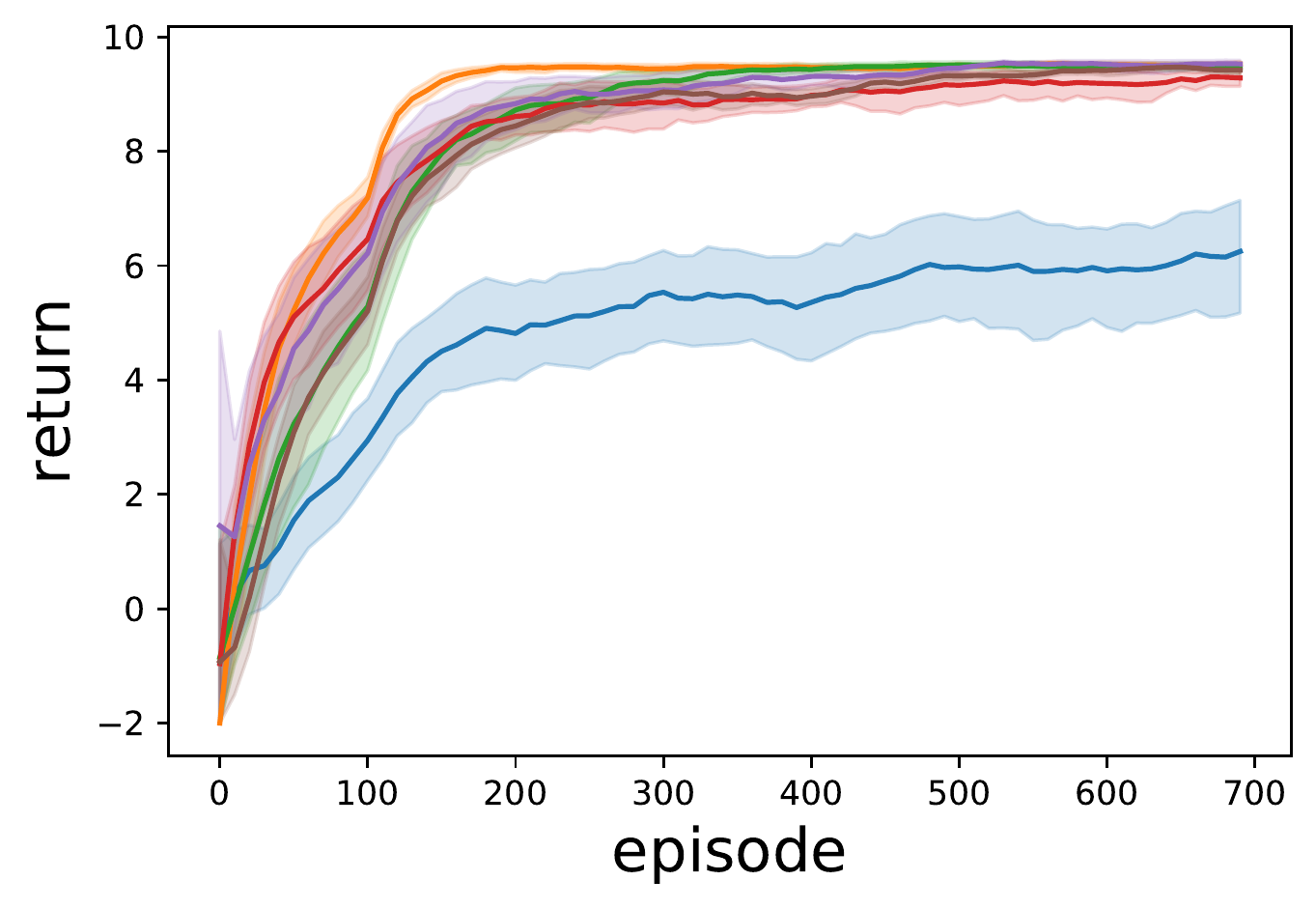}
          }
          \subfigure[transfer from 3-step gridworld]{
               \includegraphics[width=0.25\textwidth]{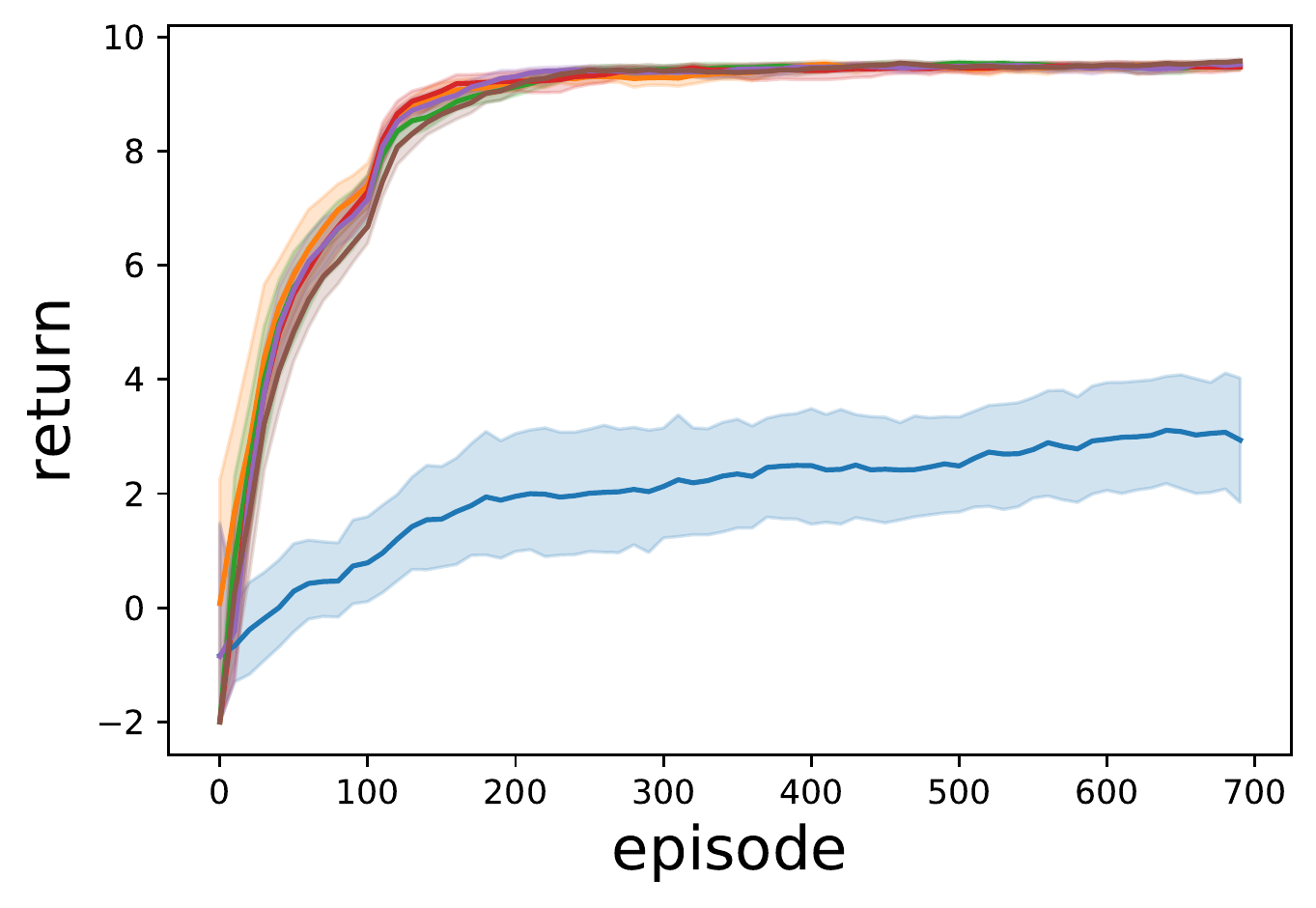}
          }
          % \subfigure[$n=3$(source task $n=1$)]{
          %     \includegraphics[width=0.3\textwidth]{fig/appendix-dim-0-2.pdf}
          % }
          \caption{The experiment results with different action embedding dimensions on 2-step gridworld.}
          \label{fig: d-transfer-result}
     \end{figure*}

     \begin{figure}
          \centering
          \subfigure[mDP (source task rP)]{
               \includegraphics[width=0.22\textwidth]{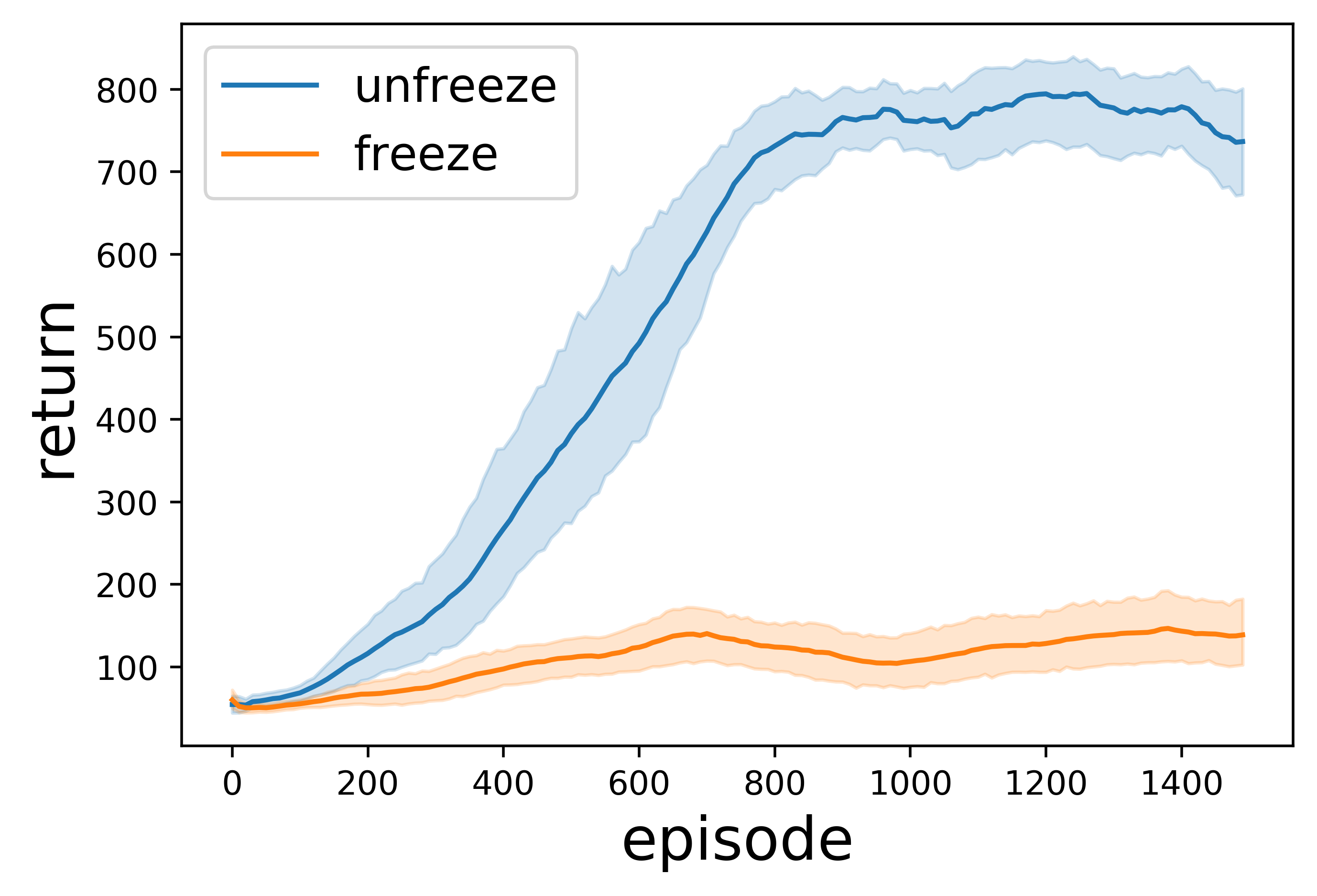}
          }
          \subfigure[rDP (source task rp)]{
               \includegraphics[width=0.22\textwidth]{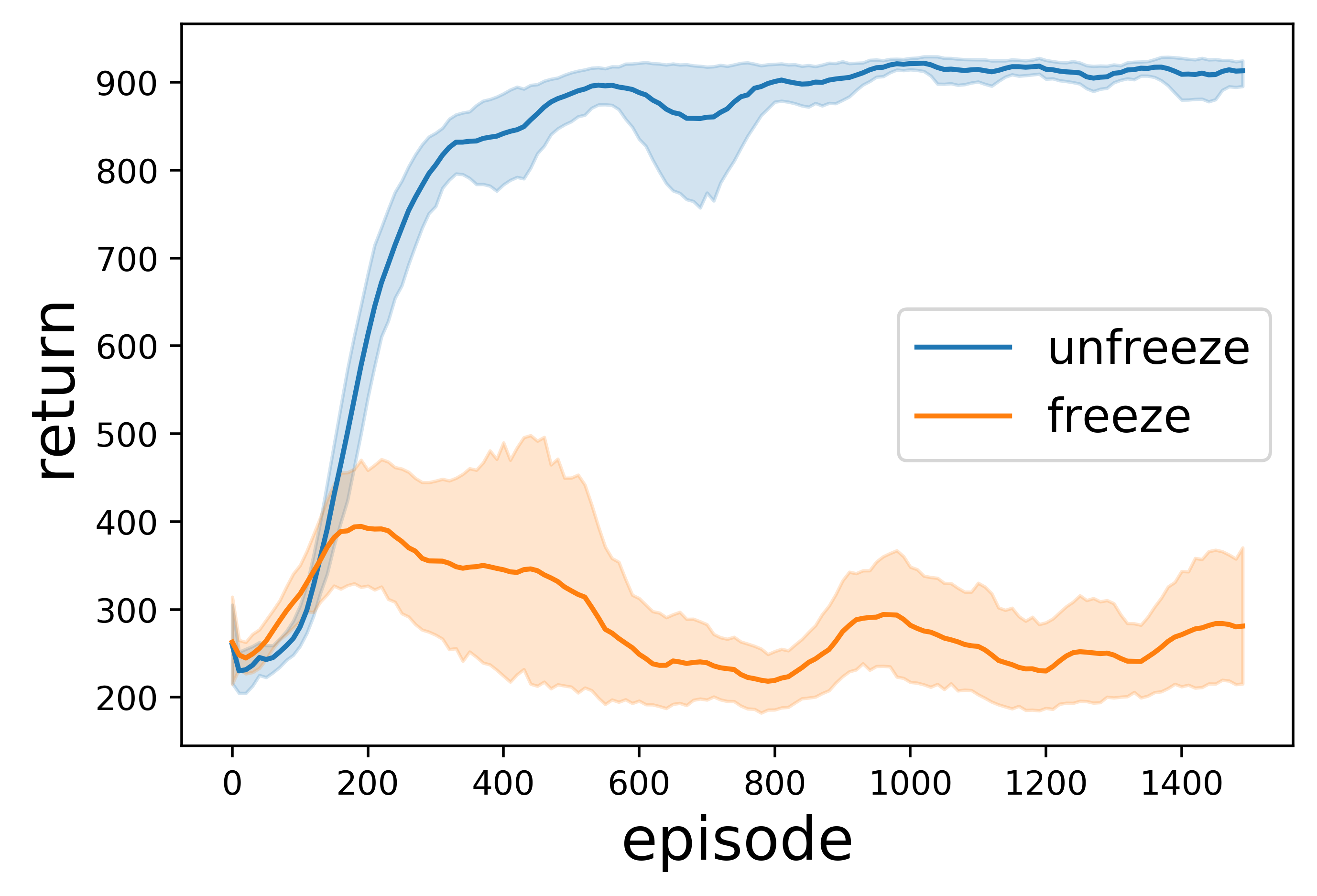}
          }
          \caption{Transfer results on mDP and rDP.}
          \label{fig: freeze}
     \end{figure}

     \section{Impact of Dimension of Action Embeddings}
     In this section, we analyze the impact of the dimension of action embeddings on the semantics of learned action embeddings and transfer performance. We conduct experiments on gridworld navigation tasks with different action embedding dimensions $d \in [1, 6]$.

     Figure \ref{fig: d-embedding} plots the visualizations of learned embeddings. Generally, as the dimension of action embeddings increases, the result appears to become more confusing because additional dimensions will introduce more noises.

     Figure \ref{fig: d-transfer-result} report the training and transfer performances with different action embedding dimensions. First, we notice that it fails to learn when $d = 1$ because it is not enough to learn action semantics.
     Figure \ref{fig: d-transfer-result}(a) shows that the training speed decreases with the increase of dimension $d$ because the action space gets larger. However, the transfer performances are only slightly influenced, as shown in Figure \ref{fig: d-transfer-result}(b) and (c). The optimal setting of $d$ will depend on the dynamics of environments and action semantics.

     \section{Freeze Parameters of the Transition Model}
     In this section, we compare the performance between TRACE whether the transition model's parameters are frozen in cross-domain transfer.
     Figure \ref{fig: freeze} depicts the result. We can see that freezing the parameters of the transition model in cross-domain transfer results in a unstable training and the negative transfer.

     % \textcolor{red}{TODO: fixed unlearned action embedding?}
\end{appendix}
\end{document}